\documentclass[10pt]{article} 
\usepackage[preprint]{tmlr}

\usepackage{amsmath,amsfonts,bm}

\def\eqref#1{equation~\ref{#1}}

\def\1{\bm{1}}

\def\rb{{\textnormal{b}}}

\DeclareMathAlphabet{\mathsfit}{\encodingdefault}{\sfdefault}{m}{sl}
\SetMathAlphabet{\mathsfit}{bold}{\encodingdefault}{\sfdefault}{bx}{n}

\usepackage{hyperref}
\usepackage{url}
\usepackage{graphicx} 
\usepackage{caption} 

\usepackage{algorithm}
\usepackage{algorithmic}

\usepackage{newfloat}
\usepackage{listings}

\usepackage{amsmath, amsfonts, bm}

\usepackage{amssymb}
\usepackage{mathtools}
\usepackage{amsthm}
\usepackage{latexsym}
\usepackage{multirow}

\usepackage{soul}
\usepackage{subcaption}
\usepackage{colortbl}
\usepackage{booktabs}
\usepackage{scalerel}
\usepackage{tablefootnote}
\usepackage{footmisc}
\usepackage{ragged2e}
\usepackage{varioref}
\usepackage{cleveref}
\usepackage{inconsolata}
\usepackage{arydshln}
\usepackage{tabularx}
\usepackage{appendix}
\usepackage{xcolor}

\newcommand{\noteng}[1]{\textcolor{red}{\bf\small NG: #1}}
\newcommand{\pg}[1]{\textcolor{blue}{\bf\small PG: #1}}
\newcommand{\abhi}[1]{\textcolor{orange}{\bf\small AN: #1}}
\newcommand{\tb}[1]{\textcolor{brown}{ #1}}

\newcommand{\fpdm}{\textbf{\emph{FastDoc}}}
\newcommand{\rbe}{$\text{RoBERTa}_{\text{BASE}}$~}
\newcommand{\bert}{$\text{BERT}_{\text{BASE}}$~}
\newcommand{\emrb}{$\text{EManuals}_{\text{RoBERTa}}$~}
\newcommand{\emrbx}{$\text{EManuals}_{\text{RoBERTa}}$}
\newcommand{\embert}{$\text{EManuals}_{\text{BERT}}$~}

\newcommand{\scifact}{\textsc{SciFact}}
\newcommand{\scibert}{\textsc{SciBERT}}
\newcommand{\sci}{\textsc{Sci}}
\newcommand{\onlybert}{\textsc{BERT}}
\newcommand{\bioroberta}{BioMedRoBERTa}

\newcommand{\componenttwo}{\textsc{RationaleSelection}}

\newcommand{\wl}{\textsc{WL}}

\title{\fpdm: Domain-Specific Fast Continual Pre-training Technique using Document-Level Metadata and Taxonomy}

\author{\name Abhilash Nandy \email nandyabhilash@kgpian.iitkgp.ac.in \\
      \addr Department of Computer Science\\
      Indian Institute of Technology Kharagpur
      \AND
      \name Manav Nitin Kapadnis \email mkapadni@andrew.cmu.edu \\
      \addr School of Computer Science\\
      Carnegie Mellon University
      \AND
      \name Sohan Patnaik \email sohanpatnaik106@iitkgp.ac.in \\
      \addr Department of Mechanical Engineering\\
      Indian Institute of Technology Kharagpur
      \AND
      \name Yash Parag Butala \email ypb@andrew.cmu.edu \\
      \addr School of Computer Science\\
      Carnegie Mellon University
      \AND
      \name Pawan Goyal \email pawang@cse.iitkgp.ac.in \\
      \addr Department of Computer Science\\
      Indian Institute of Technology Kharagpur
      \AND
      \name Niloy Ganguly \email niloy@cse.iitkgp.ac.in \\
      \addr Department of Computer Science\\
      Indian Institute of Technology Kharagpur}

\def\month{05}  
\def\year{2024} 
\def\openreview{\url{https://openreview.net/forum?id=RA4yRhjoXw}} 

\begin{document}

\maketitle

\begin{abstract}
In this paper, we propose \fpdm\ (\textbf{Fast} Continual Pre-training Technique using \textbf{Doc}ument Level Metadata and Taxonomy), a novel, compute-efficient framework that utilizes Document metadata and Domain-Specific Taxonomy as supervision signals to continually pre-train transformer encoder on a domain-specific corpus. The main innovation is that during domain-specific pretraining, an open-domain encoder is continually pre-trained using sentence-level embeddings as inputs (to accommodate long documents), however, fine-tuning is done with token-level embeddings as inputs to this encoder.  We perform 
 such domain-specific pre-training on three different domains namely customer support, scientific, and legal domains, and compare performance on 
6 different downstream tasks and 9 different datasets. 
The novel use of document-level supervision along with sentence-level embedding input for pre-training reduces pre-training compute by around $1,000$, $4,500$, and $500$ times compared to MLM and/or NSP in Customer Support, Scientific, and Legal Domains, respectively\footnote{Code and datasets are available at \url{https://github.com/manavkapadnis/FastDoc-Fast-Pre-training-Technique/}}.
The reduced training time does not lead to a deterioration in performance. In fact we show that \fpdm~either outperforms or performs on par with several competitive transformer-based baselines in terms of character-level F1 scores and other automated metrics in the Customer Support, Scientific, and Legal Domains.  Moreover, reduced training aids in mitigating the risk of catastrophic forgetting. Thus, unlike baselines, \fpdm~shows a negligible drop in performance on open domain. 
\end{abstract}
\section{Introduction}






In present times, continual  pre-training \citep{empmultidomain, dontstop} on unlabelled, domain-specific text corpora (such as PubMed articles in medical domain, research papers in Scientific Domain, E-Manuals in Customer Support Domain, etc.) has emerged as an important training strategy in NLP to enable open-domain transformer-based language models 
perform various downstream NLP tasks such as Question Answering (QA), Named Entity Recognition (NER), Natural Language Inference (NLI), etc. on domain-specific datasets \citep{cuad, scibert, nandy-etal-2021-question-answering}.  Most of the pre-training strategies involve variants of Masked Language Modelling (MLM) \citep{roberta}, Next Sentence Prediction (NSP) \citep{bert}, Sentence Order Prediction (SOP) \citep{albert}, etc. that use local sentence/span-level contexts as supervision signals. However, such methods require a lot of pre-training data and compute. For instance -  pre-training of \bert architecture on a $3.17$ billion word corpus was performed on $8$ GPUs for around $40$ days to obtain \scibert ~\citep{scibert}.


MLM-style domain-specific pre-training makes an implicit assumption that the constituent documents are independent of each other, which may not be true always. 
Documents from a particular domain (e.g., customer support, scientific papers, legal proceedings, etc.) may be categorized into different groups by experts in that area, each group containing similar documents. This information is generally stored as either `metadata' of the document \citep{metadata1, metadata2, metadata3}, or in terms of a `taxonomy' \citep{taxonomy1, taxonomy2} of documents.
For example, E-manuals of different versions of a cell phone series are very similar, 
scientific articles written on a particular topic (e.g., pre-training) follow a certain type of taxonomy, legal proceedings on related crimes are similar. 
While few models such as LinkBERT \citep{linkbert}, MetricBERT \citep{metricbert}, etc.  have used 
document metadata as an additional signal, 
no work to the best of our knowledge has \textit{singularly} leveraged taxonomy-based information \footnote{Detailed Prior Art is described \emph{\textbf{in Section \ref{sec:priorart} of Appendix.}}}.


Contrarily, in this paper, we completely replace the local context-based supervision (MLM, NSP, etc.) during pre-training with (a). \ul{document similarity} learning task using the available domain-specific metadata (through a triplet network), and (b).  \ul{hierarchical classification} task that predicts the hierarchical categories corresponding to the domain-specific taxonomy in a supervised manner.  

\begin{figure*}[!thb]
    \centering
    \includegraphics[width=\textwidth]{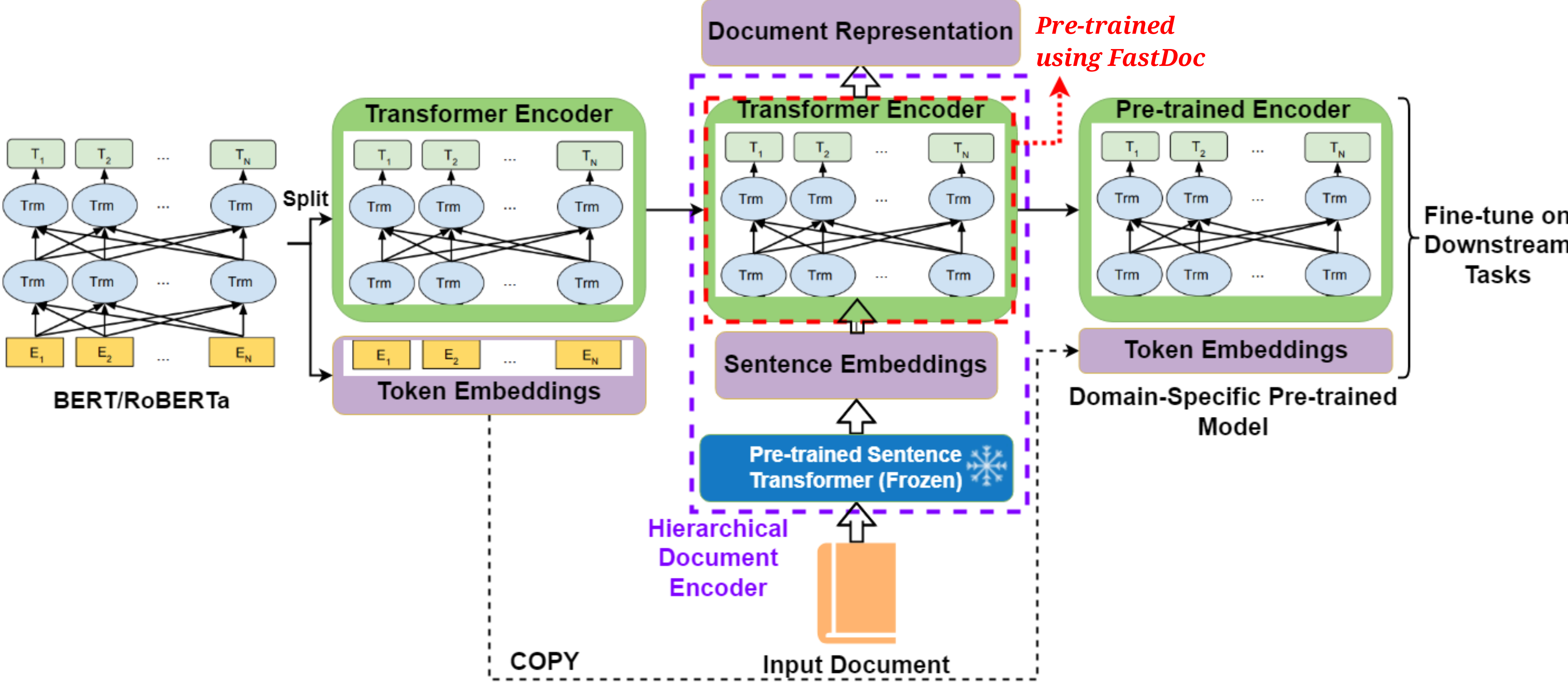}
    \caption{End-to-end training pipeline using \fpdm}
    \label{fig:pipeline}
\end{figure*}

However, to leverage document-level supervision, a robust encoding of documents is required. We use a hierarchical architecture \citep{hibert} and propose various innovations (see Figure \ref{fig:pipeline}) - (a). We initialize the lower-level encoder using a pre-trained sentence transformer (sBERT/sRoBERTa \citep{sbert}) and freeze its weights. We then initialize the higher-level encoder using pre-trained BERT/RoBERTa encoder, which now operates with a sentence embedding input, received via the lower-level encoder. This design choice (inspired by works that initialize a larger encoder through a smaller pre-trained encoder - e.g., Bert2BERT \citep{bert2bert}) helps us to directly work with sentence embeddings as inputs which in turn enables much larger contexts in a single input, and decreases the required pre-training compute by a huge margin. (b). After pre-training,
we use \ul{only the higher-level encoder} for downstream sentence and token-level tasks. As the higher-level encoder was originally pre-trained with token embedding inputs, it can still be fine-tuned with token embedding inputs. We conduct various experiments to analyze this very interesting and surprising aspect of interoperability of token and sentence embedding inputs.


Using these  ideas, we propose \fpdm\
pre-training framework, and apply it to varied NLP tasks across three disparate domains - \textbf{Customer Support}, \textbf{Scientific Papers}, and \textbf{Legal Domain}, to evaluate the generalizability of \fpdm\ across multiple domains\footnote{Continually Pre-training a single model across domains does not give good performance in all domains. That is why there are works for developing models for a particular domain, such as BioBERT \cite{biobert}, SciBERT \cite{scibert}, EManuals-BERT \cite{nandy-etal-2021-question-answering}, Legal-BERT \cite{legalbert}, FinBERT \cite{finbert}}. 
Customer Support requires answering consumer queries related to device maintenance, troubleshooting, etc., and hence, we apply \fpdm\ on two \textbf{Question Answering} tasks. In the domain of scientific papers, we focus on tasks such as extracting important scientific keywords 
\citep{bc5cdr,jnlpba,ncbi}, extracting the type of relation between such keywords 
\citep{chemprot, scierc}, as well as  classifying citation intents \citep{scicite}. In the legal domain, we focus on the task of automating \textit{contract review} 
\citep{cuad}, which involves finding key clauses in legal contracts.




{We show that \fpdm~\ul{drastically reduces} (order of 500x) pre-training compute across domains 
while still achieving comparable to  modestly better performance in downstream tasks. 
We further show that the result holds even when we increase model size and consider situations where document metadata and taxonomy may not be explicitly available. 
We also show that the frugal pre-training helps \fpdm\ 
resist catastrophic forgetting so very common when transformers undergo continual in-domain pre-training \citep{dontstop,empmultidomain}.}

\section{\fpdm\ Framework}
\label{section:pretrain}

The aim of \fpdm~is to learn robust representations for documents (in specialized domains) using potent document-level supervision signals.  
We treat a document as a sequence of sentences and provide pre-trained sentence embeddings as input. 
This, in turn, helps in accommodating documents that contain more than 512 tokens even using a standard \bert
/\rbe encoder (e.g. from Figure \ref{fig:appendix_token_distro} \emph{\textbf{in Section \ref{pre_train_suppl} of Appendix}}, we observe that using sentences as inputs enable coverage of around $50\%$ more documents than when tokens are inputs).  
We train the network with two losses. (a). {\sl The first loss} is a contrastive or triplet loss based on the similarity or dissimilarity of a document with a pair of documents; 
(b). {\sl The second loss} is a supervised loss derived while classifying a document to a domain-specific taxonomy.  


Figure \ref{fig:pipeline} depicts the end-to-end training pipeline using the proposed \fpdm\ architecture (detailed pre-training architecture shown in Figure \ref{fig:arch} \emph{\textbf{in Section \ref{pre_train_suppl} of Appendix}}). Typically a {hierarchical document encoder} like HiBERT \citep{hibert} would be a suitable model for encoding documents. It has a lower-level encoder with token inputs and a higher-level encoder with sentence-level inputs. In general, during pre-training, both these encoders need to be tuned (which is computationally expensive) and only the lower-level encoder is utilized for downstream sentence and token-level tasks such as QA, Relation Classification, NER, etc. However, we propose a different, compute-efficient method. The steps in our pipeline are - (a). The pipeline starts using an open-domain pre-trained transformer model (e.g. BERT \citep{bert}/RoBERTa \citep{roberta}) for fast convergence in domain-specific scenarios. (b). Its transformer layers excluding the input token embedding layer are used to initialize the higher-level encoder, while the lower-level encoder is a frozen sBERT/sRoBERTa. The Document representation from this document encoder is obtained by averaging the output context-aware sentence representations from the higher-level encoder. 
(c). The higher-level encoder is (further) pre-trained with document-level supervision using the proposed \fpdm\ Framework on domain-specific documents. (d). Finally, only this higher-level encoder is fine-tuned on downstream tasks, with input token embeddings copied from the open-domain model. 


Our specific design choices help in the following manner - (a) Freezing the sentence embeddings while training the encoder with document-level loss helps in achieving fast pre-training, (b) While a hierarchical encoder could also have used the document-level loss, the lower-level encoder using token inputs would be directly used for fine-tuning, but this encoder would learn less robust pre-training task-specific, semantic features as compared to the higher-level encoder \citep{prob_bert, probe_qa}. Our design trains the higher-level encoder to make the best use of pre-training loss. Next, we describe the pre-training loss functions in great detail.
The inter-operability of input token and sentence embeddings is reasoned via several experiments in Section \ref{sec:whyfpdm} and \textbf{\emph{Section \ref{FT_supp} of Appendix.}} 

\subsubsection*{\bf Contrastive Learning using document similarity labels.}
We use a Triplet Network \citep{specter}, where three documents serve as input for three document encoders, the first (anchor) and second (positive) documents being similar, and the first and third (negative) documents being dissimilar (based on metadata). 
 The encoders have hard parameter sharing~\citep{hard_parameter_sharing}. The three encoded representations are used to formulate a triplet margin loss function, denoted by $\mathcal{L}_{t}$. Mathematically,
        \begin{equation}
            \text{ $\mathcal{L}_{t}(D_1, D_2, D_3) = max\{d(D_1, D_2) - d(D_1, D_3) + 1, 0\}$}
            \label{eq:docSim}
        \end{equation}
        where $D_1, D_2, D_3$ refer to the document representations of 
        documents, and $d(.,.)$ represents the L2 norm distance. We use a unit margin in accordance with prior art using the same or similar contrastive loss functions \citep{margin1, margin2}.

Note that we do not use NT-Xent (normalized temperature-scaled cross entropy) Loss Function \citep{ntxent}, which uses multiple negatives for a given (anchor, positive) pair, as using such a large number of negatives would significantly increase the compute (corresponding to the augmentation, forward pass, and backpropagation for a large number of inputs), which defeats \fpdm’s purpose.


\subsubsection*{\bf Hierarchical Classification using Hierarchical Labels. }
Here we try to formulate a Supervised Hierarchical Classification Task based on a domain-specific hierarchical taxonomy. Given a document, the task is to predict the hierarchical categories present in the taxonomy.



In \fpdm, each document's representation
is passed through $H$ classification heads, $H$ being the maximum number of hierarchical levels present in the taxonomy. It may so happen that the hierarchy for a document has less than  $H$ levels. Hence, to  bring uniformity, a `null' class is added to each remaining level. For Hierarchical Classification, Local Classifier per Level (LCL) \citep{hierclassfn} is used, where one multi-class classifier is trained for each level of hierarchy. At each level, a classification head is an MLP layer (followed by SoftMax).
The hierarchical loss function $\mathcal{L}_{hier}$ is the sum of the categorical cross-entropy loss ($CELoss$) over all the $H$ classification heads, for all the $N$ input documents per training sample (in our case, $N = 3$). Mathematically,

{{
\begin{equation}
     \mathcal{L}_{hier} = \sum_{i=1}^{N}\sum_{j=1}^{H}CELoss(x_{ij}, y_{ij}),
\end{equation}
}}
$x_{ij}$ and $y_{ij}$ are  predicted and target class distributions 
respectively, for the $i_{th}$ document, and $j_{th}$ classification head. 

The loss $\mathcal{L}$ 
backpropagated during pre-training is the sum of the triplet margin loss and the hierarchical loss functions. 

\section{Pre-training Setup}
\label{sec:pretr_setup}
We represent BERT-based and RoBERTa-based \fpdm\ as $\fpdm_{BERT}$ and $\fpdm_{RoBERTa}$ respectively, along with abbreviation of the domain (Customer Support - \textit{Cus.}, Scientific Domain - \textit{Sci.}, Legal Domain - \textit{Leg.}). The proposed models and domain-specific baselines are pre-trained (in-domain) for $1$ epoch. We use a batch size of $32$, and AdamW optimizer \citep{adamw} with an initial learning rate of $5 \times 10^{-5}$, which linearly decays to $0$. 

We next outline the specifics of the dataset used, its associated taxonomy, metadata leveraged. 
Table \ref{tab:examples_3_domains} shows examples of  sample triplets and hierarchies from each domain.

\if{0}
\begin{table}[!thb]
\centering
\resizebox{0.9\columnwidth}{!}{%
\begin{tabular}{c|c|c}
\hline
\textbf{\begin{tabular}[c]{@{}c@{}}Domain\\ and Data\\ Source\end{tabular}} &
  \textbf{\begin{tabular}[c]{@{}c@{}}Example\\ Triplet\end{tabular}} &
  \textbf{\begin{tabular}[c]{@{}c@{}}Example\\ Hierarchy\end{tabular}} \\ \hline
\textbf{\begin{tabular}[c]{@{}c@{}}Customer\\ Support\\ (E-Manuals\\ Corpus)\end{tabular}} &
  \begin{tabular}[c]{@{}c@{}}(stereo equalizer E-Manual,\\ \ul{stereo equalizer E-Manual,}\\ \textit{blu-ray player E-Manual})\end{tabular} &
  \begin{tabular}[c]{@{}c@{}}Stereo Equalizer - \\ Electronics \textgreater \\ Audio \textgreater Audio \\ Players \& Recorders \\ \textgreater Stereo Systems\end{tabular} \\ \hline
\textbf{\begin{tabular}[c]{@{}c@{}}Scientific\\ Domain\\ (ArXiv)\end{tabular}} &
  \begin{tabular}[c]{@{}c@{}}(Proximal Policy \\ Optimization Algorithms,\\ Generating Natural \\ Adversarial Examples,\\ Autonomous Tracking of \\ Intermittent RF Source \\ Using a UAV Swarm)\end{tabular} &
  \begin{tabular}[c]{@{}c@{}}Generating Natural \\ Adversarial Examples - \\ Computer Science \textgreater \\ Machine Learning\end{tabular} \\ \hline
\textbf{\begin{tabular}[c]{@{}c@{}}Legal\\ Domain\\ (EURLEX57k)\end{tabular}} &
  \begin{tabular}[c]{@{}c@{}}("....import licences....\\ dairy products",\\ "....market research measures\\ ....milk and milk products",\\ "....importations of fishery \\ and aquaculture products....")\end{tabular} &
  \begin{tabular}[c]{@{}c@{}}"....importation of olive\\ oil...." - Agriculture \textgreater \\ Products subject to \\ market organisation \textgreater\\ Oils and fats\end{tabular} \\ \hline
\end{tabular}%
}
\caption{Examples of triplets and hierarchies in the three domains. (When representing triplets, we specify the product category in Customer Support, the paper title in the Scientific Domain, and certain key phrases from each document in the Legal Domain)}
\label{tab:examples_3_domains}
\end{table}
\fi
\begin{table*}[!thb]
\small
\centering
\resizebox{\columnwidth}{!}{%
\begin{tabular}{p{0.2\columnwidth}|p{0.5\columnwidth}|p{0.4\columnwidth}}
\hline
\textbf{Domain, Data Source} &
  \Centering\textbf{Example Triplet} &
  \Centering\textbf{Example Hierarchy} \\ \hline
  Customer Support (E-Manuals Corpus)  &
  stereo equalizer E-Manual,\newline \ul{stereo equalizer E-Manual (of a different brand),}\newline \textit{blu-ray player E-Manual}&
   \textbf{Stereo Equalizer}\newline {Electronics $\rightarrow$ Audio $\rightarrow$ Audio Players \& Recorders  $\rightarrow$ Stereo Systems} \\ \hline
   Scientific Domain (ArXiv) &
  Proximal Policy Optimization Algorithms \newline \ul{Generating Natural Adversarial Examples}\newline \textit{Autonomous Tracking of RF Source  Using a UAV Swarm} &
  \textbf{Generating Natural Adversarial Examples} \newline  {Computer Science $\rightarrow$ Machine Learning} \\ \hline
{Legal Domain\newline (EURLEX57k)} &
  ``$\cdots$import licences $\cdots$ dairy products''\newline \ul{``$\cdots$ market research measures $\cdots$ milk and milk products''} \newline \textit{`` $\cdots$ importations of fishery and aquaculture products''} &
  \textbf{``$\cdots$ importation of olive oil $\cdots$''} \newline Agriculture $\rightarrow$ Products subject to market organisation $\rightarrow$ Oils and fats \\ \hline

\end{tabular}%
}
\caption{Examples of triplets and hierarchies in the 3 domains. (When representing triplets, we specify domain-specific metadata - product category in Customer Support, paper title in the Scientific Domain, and certain key phrases from each document in the Legal Domain) [2nd column] Underlined phrases denote ``positive'' documents; italicized phrases denote ``negative'' documents.}
\label{tab:examples_3_domains}
\end{table*}

\subsection{Pre-training in the Customer Support Domain}
\label{cust_supp_pretrain}

\textbf{Dataset and Triplets Chosen.}
We pre-train \fpdm\ on a subset of the E-Manuals Corpus \citep{nandy-etal-2021-question-answering} - we sample $2,000$ E-Manual triplets, such that, the anchor and positive E-Manuals 
belong to the same product category and the anchor and negative E-Manuals belong to different product categories. The amount of data is a mere $3\%$ of the entire E-Manuals Corpus. 

\noindent\textbf{Hierarchy considered.}
Google Product Taxonomy (GPrT)\footnote{\url{https://support.google.com/merchants/answer/6324436?hl=en}} ($5,583$ possible hierarchies across $7$ levels of hierarchy) is used to obtain hierarchical classification labels using (a single) category of an E-Manual.
This allows similar E-manuals (e.g. `TV' and `Monitor') to have more similar hierarchies compared to dissimilar E-Manuals (e.g. `TV' and `Refrigerator'). 
Details on mapping product category to hierarchy are mentioned \textbf{\textit{in Section \ref{pre_train_cust_supp} of Appendix.}}



\subsection{Pre-training in the Scientific Domain}
\label{scientific_pretrain}

\textbf{Dataset and Triplets Chosen.}
We pre-train \fpdm~on a subset of the ArXiv - 
we sample $2,000$ triplets of scientific papers 
based on the ``primary category'' assigned to the paper, such that, the anchor and positive papers belong to the same category, and the anchor and negative papers belong to different categories. 
For each such triplet, we add another triplet, {where the positive and anchor samples are swapped}. 
The amount of data used is negligible compared to the $1.14M$ Papers used by \scibert\ \citep{scibert} during its pre-training.
Note that several recent works have used citations as a similarity signal \citep{specter,impr_specter,linkbert}. However, a paper might cite another paper that is not similar in terms of the content. Instead, similarity based on ``primary category'' would more intuitively lead to content-based similarity.  

\noindent\textbf{Hierarchy Considered.}
ArXiv Category Taxonomy\footnote{\url{https://arxiv.org/category_taxonomy}} (consisting of $155$ possible hierarchies across $3$ levels of hierarchy) is used to obtain hierarchical classification labels for each document, where each document is already mapped to its corresponding hierarchy via the taxonomy.


\subsection{Pre-training in the Legal Domain}
\label{pretrain_legal}
\textbf{Triplets Chosen.}
We pre-train \fpdm\ on a subset of the EURLEX57K dataset \citep{eurlex} of legislative documents - we sample $2,000$ document triplets based on the list of annotated EUROVOC Concepts\footnote{\url{http://eurovoc.europa.eu/}} assigned to each document, such that, the anchor and positive documents have at least $1$ Concept in common, and the anchor and negative documents have no Concepts in common. We double the number of triplets in a way similar to Scientific Domain.  The amount of data used is negligible compared to the $8$GB of legal contracts used for domain-specific pre-training in \citet{cuad}.

\noindent\textbf{Hierarchy Considered.}
The hierarchical class assignments of the documents in the EUR-Lex Dataset \citep{eurlex_darmstadt} 
(consisting of $343$ possible hierarchies across $4$ levels of hierarchy) are used as hierarchical classification labels, where each document is already mapped to its corresponding hierarchy.

\section{Downstream Datasets/Tasks}
\label{downstream}

\if{0}


\noteng{From previous section, context-encoder (although document but it can be used for tokens). We would like to leverage this encoder for the specific downstream tasks.We perform the task of QA on datasets of customer support. So strategy to fine-tune the network with annotated qA. Two strategies -- one is open-domain  which would work for everything SQUAD is a dataset which is a generic dataset, dataspecific Fine tuning. }

\fi

 The efficacy of the pre-training framework is tested through its performance in downstream tasks. 
 We describe those tasks and the corresponding datasets used (The names of all tasks, their corresponding datasets, and domains are listed \emph{\textbf{in Table \ref{tab:appendix_datasets} of Section \ref{appendix:downstream} of Appendix}}).   
 

\subsection{Customer Support}
\label{cust_supp_datasets}

We evaluate \ul{Question Answering} Task on two datasets - single span QA on TechQA Dataset 
and multi-span QA on S10 QA Dataset (described \emph{\textbf{in Section \ref{appendix:s10_desc} of Appendix}}).  

\subsubsubsection{\bf TechQA Dataset.} TechQA~\citep{techqa} is a span-based QA dataset with questions from a technical discussion forum and the answers annotated using IBM Technotes, which are documents released to resolve specific issues. The dataset has $600$ training, $310$ dev, and $490$ evaluation QA pairs. Each QA pair is provided with the document that contains the answer, along with $50$ candidate Technotes retrieved using Elasticsearch\footnote{\url{https://www.elastic.co/products/elasticsearch}}.

\subsubsubsection{\bf Fine-tuning Setup.}
The fine-tuning is carried out in two stages - first on the SQuAD 2.0 Dataset (inspired by \citet{techqa}), 
 and then on task-specific QA datasets, which is discussed \emph{\textbf{in Section \ref{cust_supp_exps_supp} of Appendix}}). Note that results without intermediate fine-tuning on SQuAD 2.0 deteriorate, as shown \textbf{\emph{in Section \ref{cust_supp_exps_supp} of Appendix}}.


\subsection{Scientific Domain}
\label{scientific_datasets}
We use multiple datasets from {\bf \scibert\ Benchmark Datasets} (mentioned in \citet{scibert}) for training and evaluation.
The following downstream tasks and corresponding datasets are used for evaluation - (1) \ul{NER (Named Entity Recognition)}: We use the {\bf BC5CDR} \citep{bc5cdr}, {\bf JNLPBA} \citep{jnlpba}, and {\bf NCBI-Disease} \citep{ncbi} NER Datasets of the Biomedical Domain. (2) \ul{REL (Relation Classification)}: This task predicts the type of relation between entities. The {\bf ChemProt Dataset} \citep{chemprot} from the Biomedical Domain and {\bf SciERC Dataset} \citep{scierc} from the Computer Science Domain are used for evaluation. (3) \ul{CLS (Text Classification):} {\bf SciCite Dataset} \citep{scicite} gathered from Multiple Domains is used. 

\subsubsubsection{\bf Fine-tuning Setup.} 
We fine-tune and evaluate on the downstream tasks mentioned above. The hyperparameters are the same as that in \citet{scibert}.

\subsection{Legal Domain}
\label{cuad_desc}

\textbf{CUAD} (Contract Understanding Atticus Dataset) \citep{cuad} is used, which is annotated by legal experts for the \ul{task of Legal Contract Review}. It consists of $13,101$ clauses across $41$ types of clauses annotated from $510$ contracts. Given a contract, for each type of clause, the task requires extracting relevant clauses as spans of text related to the clause type. Details of the dataset splits are given \textbf{\emph{in Section \ref{appendix:legal_downstream} of Appendix.}}

\subsubsubsection{\bf Fine-tuning Setup.}
We fine-tune and evaluate on the Contract Review Task on CUAD. The hyperparameters are the same as that in \citet{cuad}.
\section{Experiments and Results}
\label{expts_results_section}

To assess the performance of our proposed methods, we fine-tune and evaluate these methods and baselines on the datasets described in Section \ref{downstream},
 and draw inferences. Due to space constraints, the performance on the S10 QA Dataset is reported \emph{\textbf{in Section \ref{cust_supp_exps_supp} of Appendix}}. 
 In all these experiments, we perform an ablation study by considering each of the two losses of \fpdm\ separately i.e. we use only Triplet Loss ($triplet$) and only Hierarchical Classification Loss ($hier.$). We perform several additional ablations (see \emph{\textbf{Section \ref{appendix:exps} of Appendix}}) - (1) Pre-training both lower and higher-level encoders (entire hierarchical architecture), followed by fine-tuning the lower encoder worsens performance, suggesting - higher-level encoder learns better task-specific features (2) replacing sBERT/sRoBERTa with BERT/RoBERTa worsens performance, suggesting - sentence transformers provide effective sentence embeddings. (3). used a more fine-grained document similarity criterion (changed Eq. \ref{eq:docSim}) and found the result to be inferior, and (4). compared \fpdm\ with the much larger GPT-3.5 model and found that GPT-3.5 models perform much inferior in 0 and 1-shot settings. 

 




\subsection{Customer Support Domain}




\if{0}
\noindent{\bf Evaluation Metrics.} For the S10 Dataset, the evaluation metrics for the downstream tasks of Section Retrieval and Answer Retrieval are as follows:
\noindent \textbf{Answer Retrieval} - 
HA\_F1@1 reported for TechQA Dataset is also reported for S10 QA Dataset. However, we do not report the F1 and HA\_F1@5
metrics for the S10 Dataset, as a single answer is retrieved per question, rather than multiple answers with corresponding scores. The other metrics reported are (a).Exact Match, which is the fraction of questions for which the retrieved answer exactly matches the ground truth answer; (b). ROUGE-L \citep{rouge}, which is an F1-measure evaluation metric calculated on the basis of the longest common subsequence between the retrieved answer and the ground truth; (c). Sentence and Word Mover Similarity \citep{swms} - GloVe word vectors~\citep{glove} are weighed by the individual word frequencies to get sentence embeddings, and sentence embeddings are weighted by the sentence lengths to get the final embedding. A Linear Programming Solution is worked out to measure the distance between the embedding vectors of the retrieved answer and the ground truth.  EM, ROUGE-L F1, and S+WMS metrics, as a considerable number of questions are unanswerable ($150$ out of $600$ questions in the training set and $150$ out of $310$ questions in the development set are unanswerable), and these metrics make sense when each question has a ground truth answer.
    
    \noindent \textbf{Section Retrieval} - HITS@K, which is equal to the percentage of the questions for which, the section containing the ground truth answer is one of the top $K$ retrieved sections. We report values for $K = 1$ and $K = 3$.
    
    
\fi


\textbf{Baselines: }We compare our pre-training approach to $3$ types of pre-training baselines described below. For the sake of completeness, we also compare with baselines using span/sentence-level supervision signals. Domain-specific Continual Pre-training is carried out  on the corpus of E-Manuals for all baselines (except \bert, \rbe, and Longformer). 
\\\noindent \textbf{(1) Pre-training using masked language modeling (MLM) and/or Next Sentence Prediction (NSP)}: We use \bert~\citep{bert}, \rbe~\citep{roberta}, $\text{LinkBERT}_{\text{BASE}}$ \cite{linkbert}, Longformer \citep{longformer}, 
\embert and \emrb \citep{nandy-etal-2021-question-answering} (domain continual pre-training of \bert \citep{bert} and \rbe \citep{roberta}, respectively, on the entire E-Manuals corpus). 
\noindent\textbf{(2) Using intra-document contrastive learning}: DeCLUTR \citep{declutr} and ConSERT \citep{consert} are the intra-document contrastive learning methods. 
\noindent\textbf{(3) Using inter-document contrastive learning}: SPECTER \citep{specter} is the inter-document contrastive learning baseline used. (Details on tailoring SPECTER to Customer Support are given \textbf{\emph{in Section \ref{cust_supp_exps_supp} of Appendix}}). 

\begin{table}[!htb]
\centering
\small
\begin{tabular}{l|ccc}
\hline
                                                                                                                                & \textbf{F1}    & \textbf{HA\_F1@1} & \textbf{HA\_F1@5} \\ \hline
 \bert       & 13.67          & 26.49             & 36.14                         \\
                                                                                                        \rbe       & 16.46          & 31.89             & 42.4
                                    \\    $\text{LinkBERT}_{\text{BASE}}$ & 14.24 & 27.59 & 36.77                       \\       Longformer & 16.57 & 32.1 & 42.66                                  \\ \embert          & 13.41          & 25.98             & 36.69                       \\
                                                                                                        \emrb       & 16.04          & 31.08             & 44.71              \\
                                                                                                        DeCLUTR                 & 15.11          & 29.28             & 38.93                       \\
                                                                                                        ConSERT                 & 11.12          & 21.54             & 30.37                       \\
                                                                                                        SPECTER                 & 12.92          & 25.03             & 34.74                       \\ 
 \fpdm$(Cus.)_{BERT}$(hier.)           & 14.19          & 27.49             & 36.62                    \\
                                                                                                        \fpdm$(Cus.)_{BERT}$(triplet)                 & 14.47          & 28.04             & 37.21                       \\
                                                                                                        \fpdm$(Cus.)_{BERT}$ & 14.56          & 28.2              & 35.54                       \\ \hline 
 \fpdm$(Cus.)_{RoBERTa}$(hier.)           & 16.52         & 32.00               & 44.77                    \\
                                                                                                        \fpdm$(Cus.)_{RoBERTa}$(triplet)                 & 16.39          & 31.76             & \textbf{46.59}             \\
                                                                                                        \fpdm$(Cus.)_{RoBERTa}$ & \textbf{17.52} & \textbf{33.94}    & 44.96                      \\ \hline 
  
\end{tabular}%
\caption{Results for the QA task on the TechQA Dataset. 
}
\label{tab:techqa_results}
\end{table}







\noindent{\bf Performance on TechQA Dataset } The answer-retrieval performance on the development set (as per \citet{techqa}) is reported in Table~\ref{tab:techqa_results}. The model gives five candidate answers per question and corresponding confidence scores. Each answer is assigned an `evaluation score' - If the confidence score is below a threshold provided by the model, `evaluation score' is $1$ if the question is actually unanswerable, and $0$ otherwise.  However, if the confidence score is above the threshold, the `evaluation score' is character F1 between the predicted answer and ground truth and $0$ if the question is actually unanswerable. 
The evaluation metrics used, as mentioned in ~\citet{techqa}\footnote{We do not use BEST\_F1, as a threshold is tuned on the dev. set using F1 score, which is not realistic}, are (a). \textbf{F1} - `evaluation score' for the predicted answer (with the highest confidence score) averaged across all questions.
    (b). \textbf{HA\_F1@1} -  similar to F1, except that, the averaging is done on the answerable question set ($160$ out of $310$ questions in the dev set are answerable).
    (c). \textbf{HA\_F1@5} - macro average of the 
    5 best candidate answers per question, averaged across the answerable question set.

From the results in Table~\ref{tab:techqa_results}, we can infer - (1). Among the baselines, 
(a) Longformer gives the best F1 and HA\_F1@1 and the second-best HA\_F1@5. This is because of the long sequence length of $4,096$ compared to $512$ of other models\footnote{For completeness, we have also continually pre-trained Longformer on the data used by \fpdm, and it shows inferior results to Longformer on 2/3 metrics due to the data being insufficient to adapt Longformer.}.
(b) SPECTER does not perform well, even though it uses document-level supervision, as it cannot accommodate the entire document within $512$ tokens, so only the first $512$ tokens are used which does not help much in learning. 
{(c) ConSERT performs contrastive learning on sentence inputs, prohibiting it from learning context beyond a single sentence (unlike \fpdm\ that learns inter-sentence context during pre-training due to its hierarchical architecture), thus reducing performance on QA tasks.}
(d) In general, contrastive learning baselines perform inferior to those using MLM/NSP. (2). $\fpdm\textit{(Cus.)}_{BERT}$ variants perform better than BERT-based baselines, 
and $\fpdm\textit{(Cus.)}_{RoBERTa}$ variants than almost all RoBERTa-based baselines,
suggesting that our proposed pre-training methods are better than that of baselines. (3) $\fpdm\textit{(Cus.)}_{RoBERTa}$ variants perform better than $\fpdm\textit{(Cus.)}_{BERT}$ variants, as RoBERTa \citep{roberta} performs better than BERT \citep{bert} in span-based QA tasks such as SQuAD \citep{squadv2, squadv1}.
(4) $\fpdm\textit{(Cus.)}_{RoBERTa}$ performs the best of all models {
in F1 and HA\_F1@1} and the second-best in HA\_F1@5. $\fpdm\textit{(Cus.)}_{RoBERTa}$ performs around $6\%$ \ul{better than the best baseline Longformer} both in terms of F1 and HA\_F1@1 (even though Longformer has a long sequence length, it is not able to encode most documents properly). 
 


Additionally, we perform a qualitative analysis \textbf{\textit{in Table \ref{tab:qual_analysis_full} of Section \ref{cust_supp_exps_supp} of Appendix}} by comparing the ground-truth answers and the answers predicted by \fpdm\ and a well-performing baseline for $3$ answerable questions in TechQA and S10 QA Datasets. This analysis suggests that \fpdm\ is comparatively better at extracting numerical entities, tackling multiple questions in a sample, and answering location-based questions. 




\subsection{Scientific Domain}
\label{sci_domain_exps}

\begin{table}[!thb]
	\centering
        \small
\begin{tabular}{lllc|ccc}
\hline
\textbf{Field}       & \textbf{Task}        & \textbf{Dataset} & \scibert & \begin{tabular}[c]{@{}c@{}}\fpdm\\\textit{\textbf{(triplet)}}\end{tabular} & \begin{tabular}[c]{@{}c@{}}\fpdm\\\textbf{\textit{(hier.)}}\end{tabular} & \fpdm \\ \hline
\multirow{4}{*}{BIO} & \multirow{3}{*}{NER} & BC5CDR           & 85.55          & 87.7 & \textbf{87.94}    & 87.81                 \\
      &     & JNLPBA & 59.5  &  75.86 & \textbf{75.97} & 75.84 \\
      &     & NCBI-D & \textbf{91.03} &   84.15 & 87.81 &  84.33 \\
      & REL & ChemProt  & 78.55  &  75.12 & 80.28 & \textbf{80.48} \\
      \hline
CS   & REL & SciERC       & 74.3  &  75.4 & 75.62 & \textbf{78.95} \\
\hline
Multi & CLS & SciCite      & 84.44  &  84.31 & \textbf{84.48} &  83.59\\
\hline
\end{tabular}

	\caption{$\fpdm(Sci.)_{BERT}$ and its variants vs. \scibert\ in tasks presented in \citet{scibert}. Following \citet{scibert}, we report macro F1 for NER (span-level), and for REL and CLS (sentence-level), except for ChemProt, where we report micro F1.}
	\label{tab:scibert_paper_tasks}
\end{table}

\noindent \textbf{Baselines: }\scibert\ \citep{scibert} (pre-trained using  MLM and NSP on a huge scientific corpus)\footnote{vocabulary used for \scibert\ is same as that of \bert for consistency among \scibert\ and \fpdm\ variants. Specifically, we use this model as \scibert\ - \url{https://s3-us-west-2.amazonaws.com/ai2-s2-research/scibert/huggingface_pytorch/scibert_basevocab_uncased.tar}}. 

\noindent \textbf{Performance on Different Datasets } We fine-tune and evaluate on the datasets mentioned in Section \ref{scientific_datasets}. The results on the test set for each task are shown in Table \ref{tab:scibert_paper_tasks}. We see that $\fpdm(Sci.)_{BERT}$ performs better than \scibert\ on $4$ out of $6$ datasets, and performs the best on the Relation Classification Tasks. However, $\fpdm(Sci.)_{BERT}(hier.)$ performs the best on $3$ datasets with NER and text classification tasks, as (1) fine-grained NER benefits from fine-grained hierarchical information, and (2) text classification dataset has samples from multiple domains, where diversity in the hierarchical categories helps. 

Since recent works have used citations as a similarity signal, we report the performance of \fpdm\ using citations as a similarity signal \emph{\textbf{in Table \ref{tab:appendix_scibert_paper_tasks} of Section \ref{appendix:scientific_exps} of the Appendix.}} This gives a satisfactory performance, showing that \fpdm\ works on different types of metadata. However, on average, a system using citations does not perform as well as when using ``primary category".

\subsection{Legal Domain}


\textbf{Baselines: } We use baselines from \citet{cuad} - \bert, \rbe, \rbe + Contracts Pre-training (domain-specific pre-training of RoBERTa-BASE on \~8GB of unlabeled contracts collected from the EDGAR database). Also, we use CDLM \cite{cdlm}, LEGAL-BERT-FP \citep{legalbert}, and LEGAL-RoBERTa-BASE \citep{legalroberta} as additional baselines.

\subsection*{\bf Performance on CUAD Dataset}

\begin{table}[!thb]
\centering
\small
\begin{tabular}{l|c|c}

\hline \textbf{Model} &
              \multicolumn{1}{c|}{\textbf{AUPR}}                        & \begin{tabular}[c]{@{}c@{}}\textbf{Precision@}\\ \textbf{80\% Recall})\end{tabular} \\ \hline

\bert & 32.4  & 8.2 \\
LEGAL-BERT-FP & 32.6 & 21.16 \\
\rbe & 42.6  & 31.1 \\
LEGAL-RoBERTa-BASE & 42.9 & 31.7 \\
\rbe + Contracts Pre-training & {\bf 45.2}  & 34.1 \\
CDLM & 43.2 & {\bf 34.6}                  \\ 
\hline
$\fpdm(Leg.)_{BERT}(triplet)$ &   32.5   & 8.3                  \\
$\fpdm(Leg.)_{BERT}(hier.)$ &   32.8   & 9.4                  \\
$\fpdm(Leg.)_{BERT}$ &   32.6   & 9.4                  \\
$\fpdm(Leg.)_{RoBERTa}(triplet)$ & 42.4  & 32.7                  \\ 
$\fpdm(Leg.)_{RoBERTa}(hier.)$ & 42  & 32.3                  \\ 
$\fpdm(Leg.)_{RoBERTa}$ & 44.8  & {\bf 34.6}                  \\ 
\hline
\end{tabular}%

\caption{$\fpdm(Leg.)$ and its variants vs. baselines in the Contract Review task on CUAD (AUPR - Area Under Precision-Recall Curve).}
\label{tab:cuad_results}
\end{table}

{CUAD exhibits class imbalance, rendering AUPR (Area Under Precision-Recall) and Precision@80\% Recall as suitable metrics. Furthermore, AUPR effectively encapsulates model performance across various confidence thresholds.}
From Table \ref{tab:cuad_results}, we infer - (1) All $\fpdm(Leg.)_{BERT}$ variants perform better than $\text{BERT}_{\text{BASE}}$, and $\fpdm(Leg.)_{RoBERTa}$ performs better than \rbe and Legal-RoBERTa-BASE (2) $\fpdm(Leg.)_{RoBERTa}$ performs better than CDLM, even though CDLM has a long sequence length of 4096 (3) $\fpdm(Leg.)_{RoBERTa}$ gives the best Precision@$80\%$ Recall, and second-best AUPR, 
even though it uses negligible domain-specific pre-training data compared to baselines.

\noindent{\textbf{Summary of the Experiments and Results.}} Note that although \fpdm\ uses document information, it does not use the information derived from MLM during continual pre-training that many other conventional domain-specific baselines use. Therefore, we maintain that our approach is both equitable and innovative when compared to the baselines. We observe that across various domains and different types of generalization, \fpdm\ typically outperforms (albeit modestly) the baselines, and in cases where it falls short, the difference is marginal. This happens despite using much less compute which is elaborated in Section \ref{pretraincompute}. 




\subsection{Utility of the pre-training losses: Examples}
\label{section:utility}

We looked into the datasets and gauged the impact of the two pre-training losses in \fpdm's performance. Specifically, we took cases where  \fpdm\ has produced better results than a well-performing baseline and chose some representative examples to present. 
Table \ref{tab:analysis_3_domains} shows examples from each domain where triplet and hierarchical losses are beneficial, along with probable reasons. We can see that domain-specific knowledge present in the metadata and taxonomy helps \fpdm\ in performing well on domain-specific downstream tasks.

\begin{table*}[!thb]
\small
\centering
\resizebox{0.9\textwidth}{!}{%

\begin{tabular}
{p{0.05\textwidth}|p{0.09\textwidth}|p{0.4\textwidth}|p{0.4\textwidth}}
\hline
 & { Dataset} & \Centering{\textbf{Triplet Loss is beneficial}} & \Centering{\textbf{Hierarchical Loss is beneficial}} \\ \hline
\textbf{Cus.} & S10 QA (QA) & Q. How can I enable the accidental touch protection ? & Q. I need the registered fingerprint list. Where can I find this?\\ \hline 
   \multicolumn{2}{c|}{Reasons} & ``accidental touch'' benefits from triplets having anchor and positive as Touch-based device E-Manuals, and negative as an E-Manual of a device without touch screen. &``fingerprint'' benefits from multiple hierarchies with ``Biometric Monitors''. \\ \hline
  \hline
\textbf{Sci.} & SciCite (Text Classification) & A primary benefit of these models is the inclusion of variability in model parameters (Parnell et al. 2010). \textbf{Output - ``background''} & The SVR can be considered as a novel training technique; the following section presents a concise introduction to the SVR [33, 35, 38]. \textbf{Output - ``background''}\\ \hline
 \multicolumn{2}{c|} {Reasons} & Using triplets with anchor and positive belonging to ``Machine Learning" help in classifying the text as ``background", as ``model parameters" is a common term used in ``Machine Learning" papers.  & Although the first sentence could lead to the inference that ``SVR" is a new method, other papers belonging to the hierarchy ``Computer Science $\rightarrow$ Machine Learning" would suggest that ``SVR" exists already.  \\ \hline
  \hline
\textbf{Leg.}      & CUAD (Clause Extraction)  & ..to make or have made the Products anywhere in the world for \textbf{import} or \textbf{sale} in the Field in the Territory in each case,.. & ..such \textbf{commercial crops} will be interplanted as \textbf{agriculture} and forestry as well as medicinal materials;.. \\ \hline
\multicolumn{2}{c|}{Reasons} & The triplet on the concepts of ``import policy" and ``sale" in the anchor and positive is beneficial. & the hierarchy ``Agriculture $\rightarrow$ Products subject to market organisation" helps here.  \\ \hline
  \hline
\end{tabular}%
}
\caption{Samples from each domain, where Triplet Loss and Hierarchical Loss are beneficial. Note that we add outputs for the SciCite Text Classification Task for more clarity.
}
\label{tab:analysis_3_domains}
\end{table*}
\section{Pre-training Compute of \fpdm~relative to the baselines}
\label{pretraincompute}

\begin{table}[!htb]
\centering
\small
\begin{tabular}{c|l|c}
\hline Domain &
              \multicolumn{1}{c|}{Model}                        & \begin{tabular}[c]{@{}c@{}}Compute (in\\ GPU-hours)\end{tabular} \\ \hline
\multirow{7}{*}{\begin{tabular}[c]{@{}c@{}}Customer\\Support\end{tabular}} & \embert                        & 576                    \\
& \emrb                     & 980                   \\
& DeCLUTR  & 370 \\
& ConSERT  & 40 \\
& SPECTER  & 600 \\
& \fpdm$(Cus.)_{BERT}$     & 0.58                  \\
& \fpdm$(Cus.)_{RoBERTa}$  & 0.75                  \\ \hline
\multirow{2}{*}{\begin{tabular}[c]{@{}c@{}}Scientific\\Domain\end{tabular}} & \scibert & 7680 \\
& \fpdm$(Sci.)_{BERT}$ & 1.7 \\
\hline
\multirow{2}{*}{\begin{tabular}[c]{@{}c@{}}Legal\\Domain\end{tabular}} & \begin{tabular}[l]{@{}c@{}}\rbe +\\Contracts Pre-training\end{tabular} & 710\\
& \fpdm$(Leg.)_{RoBERTa}$ & 1.49 \\
\hline
\end{tabular}%

\caption{Pre-training Compute of \fpdm~vs. baselines}
\label{tab:compute}
\end{table}

We compare the compute (in terms of GPU-hours - GPUs needed multiplied by the number of hours) for pre-training \fpdm~with baselines.  
NVIDIA GeForce GTX 1080 Ti GPUs are used for pre-training.

Baselines in Customer Support and Legal Domains are continually pre-trained on the domain, while the \scibert\ baseline in Scientific Domain is pre-trained from scratch.  

\noindent \textbf{Customer Support: }Table \ref{tab:compute} shows that \fpdm$(Cus.)_{BERT}$ and \fpdm$(Cus.)_{RoBERTa}$ use roughly 1,000 times\footnote{\fpdm$(Cus.)_{BERT}$ uses 33.3x less documents compared to \embert during pre-training. Additionally, \fpdm$(Cus.)_{BERT}$ takes sentence embeddings as inputs, while \embert\ takes in token embeddings. There are 37.3 tokens per sentence in the pre-training corpus, meaning that there are 37.3x lesser samples for \fpdm$(Cus.)_{BERT}$ w.r,t \embert\ for the same text, reducing compute further from 33.3x to 1000x (33.3 $\times$ 37 is $\approx$ 1000)} and 1,300 times less compute compared to \embert and \emrb respectively, and require significantly less compute than all the baselines. 
{\sl It actually takes less than 1 GPU-hour for {pre-}training} \fpdm.  ConSERT is the closest baseline in terms of compute time, as its inputs are a limited number of sentence pairs, unlike a huge number of spans in DeCLUTR, a large number of triplets in SPECTER, and several masked sentences in \embert and \emrb. 
\noindent \textbf{Legal Domain: } \fpdm$(Leg.)_{RoBERTa}$ needs around $480$ times less compute than continual pre-training of RoBERTa-BASE on  contracts as in \citet{cuad}.

\noindent \textbf{Scientific Domain: } \fpdm$(Sci.)_{BERT}$ needs around $4,520$ times less compute than \scibert\footnote{According to \citet{scibert}, it takes a minimum of $40$ days on $8$ GPUs (elaborated \emph{\textbf{in Section \ref{appendix:pretr_compute} of Appendix}})}. The decrease in compute of \fpdm\ compared to the domain-specific baselines is much more compared to Customer Support and Legal Domains, as \scibert\ is pre-trained from scratch. \fpdm$(Sci.)_{BERT}$ continually pre-trains \bert\ on Scientific Domain, and its downstream task performance and domain-specific compute compared to \scibert\ shows that \textit{domain-specific pre-training from scratch is not necessary.}  
 
Thus, these experiments demonstrate the remarkable efficiency of the proposed pre-training paradigm, as well as the choice of the pre-training architecture used in \fpdm.
\section{Analysis and Ablations}

In this section, we report the following analysis and ablations - (1) Catastrophic Forgetting when evaluating \fpdm\ in open-domain (2) absence of document supervision in the domain of interest (3) Effect of using a larger backbone model for \fpdm\ (4) Reasons behind \fpdm\ working the way it does. Also, we apply Parameter-Efficient training on \fpdm\ as an ablation \emph{\textbf{in Section \ref{appendix:peft} of Appendix}}, which gives very poor downstream task results and is not beneficial from a compute perspective as well.



\begin{table*}[!thb]
\centering
\resizebox{\textwidth}{!}{%
\begin{tabular}{l|c|c|cc|cc|cc|c|c|c}
\hline
\multicolumn{1}{c|}{\textbf{TASK}}   & \textbf{CoLA}                         & \textbf{SST2}                                 & \multicolumn{2}{c|}{\textbf{MRPC}}                                                           & \multicolumn{2}{c|}{\textbf{STS}}                                                   & \multicolumn{2}{c|}{\textbf{QQP}}                                           & \textbf{MNLI}                        & \textbf{QNLI}                                 & \textbf{RTE}                         \\ \hline
\multicolumn{1}{c|}{\textbf{METRIC}} & \begin{tabular}[c]{@{}c@{}}\textbf{Matthews}\\\textbf{CC}\end{tabular}           & \textbf{Acc.}                             & \textbf{F1-score}                            & \textbf{Acc.}                             & \begin{tabular}[c]{@{}c@{}}\textbf{Pearson}\\\textbf{CC}\end{tabular}                    & \begin{tabular}[c]{@{}c@{}}\textbf{Spearman}\\\textbf{CC}\end{tabular}         & \textbf{F1-score}                    & \textbf{Acc.}                    & \textbf{Acc.}                    & \textbf{Acc.}                             & \textbf{Acc.}                    \\ \hline
$RoBERTa_{BASE}$                     & \textbf{63.71}                        & 94.15                                         & 92.71                                        & 89.71                                         & 90.91                                         & \textbf{90.66}                      & \textbf{89.1}                        & \textbf{91.84}                       & \textbf{87.24}                       & 92.26                                         & \textbf{80.14}                       \\ \hdashline
\fpdm$(Cus.)_{RoBERTa}$                 &  \begin{tabular}[c]{@{}c@{}}62.57\\{\color[HTML]{FD6864}(-1.14)}\end{tabular}  &  \begin{tabular}[c]{@{}c@{}}\textbf{94.27}\\{\color[HTML]{6aad77}\textbf{(+0.12)}}\end{tabular} &  \begin{tabular}[c]{@{}c@{}}\textbf{93.1}\\{\color[HTML]{579c64}\textbf{(+0.39)}}\end{tabular} &  \begin{tabular}[c]{@{}c@{}}\textbf{90.44}\\{\color[HTML]{42854f}\textbf{(+0.73)}}\end{tabular} &  \begin{tabular}[c]{@{}c@{}}\textbf{90.98}\\{\color[HTML]{80c28d}\textbf{(+0.07)}}\end{tabular} & \begin{tabular}[c]{@{}c@{}}\textbf{90.66 }\\\textbf{(0)}\end{tabular}                  &  \begin{tabular}[c]{@{}c@{}}89.08\\{\color[HTML]{FF918E}(-0.02)}\end{tabular} & \begin{tabular}[c]{@{}c@{}}\textbf{91.84}\\\textbf{(0)}\end{tabular}                   &  \begin{tabular}[c]{@{}c@{}}87.22\\{\color[HTML]{FF918E}(-0.02)}\end{tabular} &  \begin{tabular}[c]{@{}c@{}}\textbf{92.62}\\{\color[HTML]{579c64}\textbf{(+0.36)}}\end{tabular} &  \begin{tabular}[c]{@{}c@{}}79.06\\{\color[HTML]{FD6864}(-1.08)}\end{tabular} \\ \hdashline
$EManuals_{RoBERTa}$                 &  \begin{tabular}[c]{@{}c@{}}51.82\\{\color[HTML]{CB0000}(-11.89)}\end{tabular} &  \begin{tabular}[c]{@{}c@{}}91.97\\{\color[HTML]{FE0000}(-2.18)}\end{tabular}          &  \begin{tabular}[c]{@{}c@{}}91.42\\{\color[HTML]{FD6864}(-1.29)}\end{tabular}         &  \begin{tabular}[c]{@{}c@{}}87.99\\{\color[HTML]{FE6864}(-1.72)}\end{tabular}          &  \begin{tabular}[c]{@{}c@{}}88.4\\{\color[HTML]{FE0000}(-2.51)}\end{tabular}           &  \begin{tabular}[c]{@{}c@{}}88.36\\{\color[HTML]{FE0000}(-2.3)}\end{tabular} &  \begin{tabular}[c]{@{}c@{}}88.65\\{\color[HTML]{FD6864}(-0.45)}\end{tabular} &  \begin{tabular}[c]{@{}c@{}}91.55\\{\color[HTML]{FD8A87}(-0.29)}\end{tabular} &  \begin{tabular}[c]{@{}c@{}}85.15\\{\color[HTML]{FE0000}(-2.09)}\end{tabular} &  \begin{tabular}[c]{@{}c@{}}91.34\\{\color[HTML]{FD6864}(-0.92)}\end{tabular}          &  \begin{tabular}[c]{@{}c@{}}70.4\\{\color[HTML]{CB0000}(-9.74)}\end{tabular}  \\ \hline
\end{tabular}%
}
\caption{Dev. set results on GLUE Benchmark (CC - Correlation Co-efficient, Acc. - Accuracy)}
\label{tab:glue_results}
\end{table*}

\subsection{Catastrophic Forgetting in open-domain} 
\label{gluedataset}


\begin{figure}[!thb]
    \centering
    \caption{Relative change (in $Log_{10}$ Scale) in the L1-norm of different types of parameters during pre-training via MLM vs. \fpdm.}
    \includegraphics[width=0.55\textwidth]{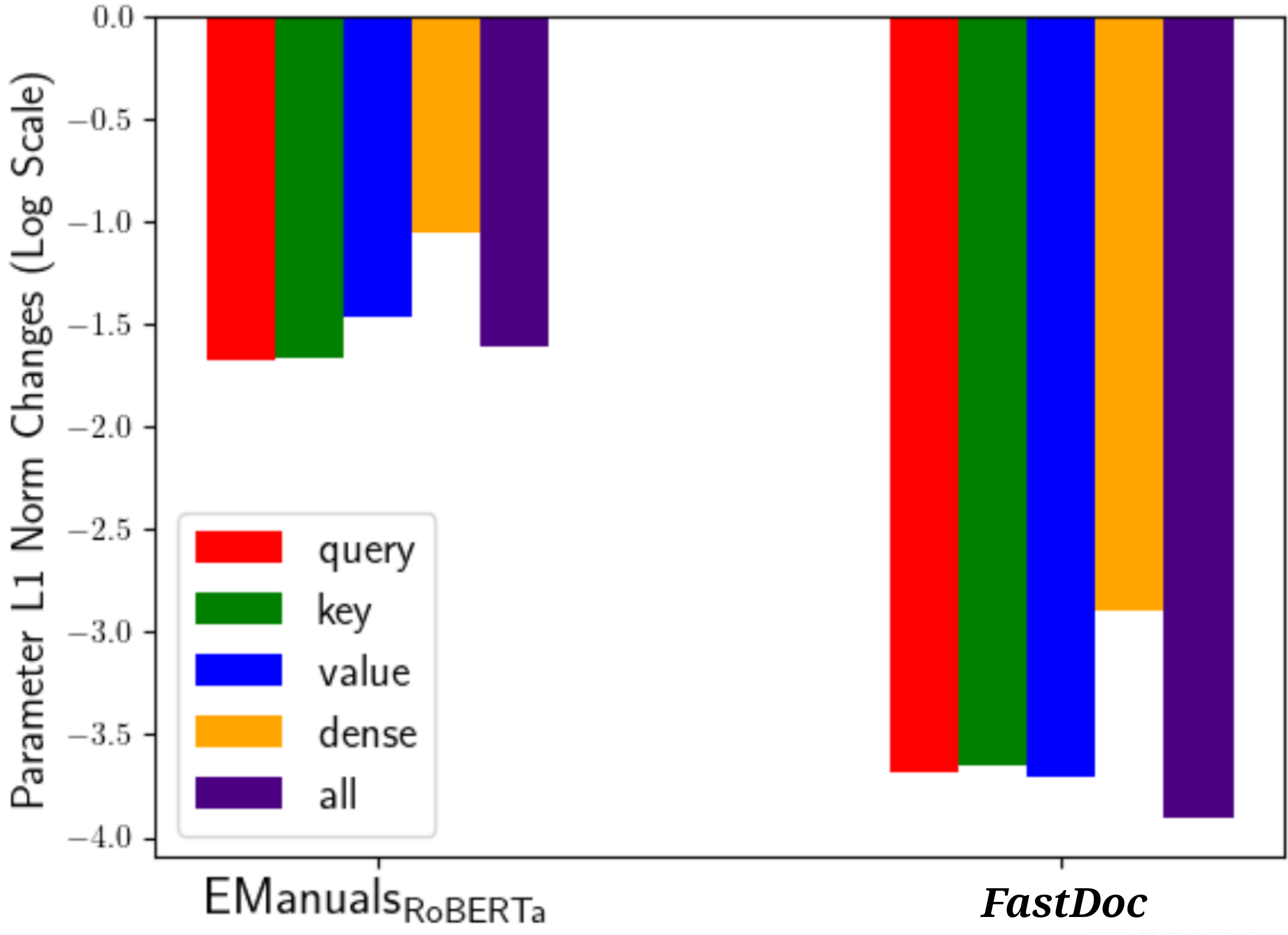}
    
    \label{fig:change_params}
\end{figure}

Recent works show that continual in-domain pre-training of transformers 
leads to a significant performance drop when fine-tuned on open-domain datasets \citep{empmultidomain,dontstop} resulting in Catastrophic Forgetting (CF). Such works start with an open-domain model (e.g. BERT/RoBERTa) and perform open-domain benchmark tasks (e.g. GLUE). Then, they consider a model pre-trained continually on a specific domain (e.g. BioBERT) and re-assess performance on those tasks. Decrease in performance of domain-specific model compared to the open-domain model determines degree of catastrophic forgetting. 

Similarly, we fine-tune \rbe (pre-trained on open-domain corpora), $\fpdm\textit{(Cus.)}_{RoBERTa}$, and \emrb from \textbf{customer support}
on the datasets of the (open-domain) GLUE \citep{glue} benchmark, and the results are shown in Table \ref{tab:glue_results}. The hyperparameters used are mentioned \emph{\textbf{in Section \ref{appendix_cf} of Appendix.}} 

We observe that - (a). $\fpdm\textit{(Cus.)}_{RoBERTa}$ performs better than \rbe in $4$ out of $8$ tasks (although the improvement is minor), even after continual pre-training on E-Manuals, while the drop in performance in the other 4 tasks is negligible. The performance improvement in tasks that require predicting relation between sentence pairs, like STS, QNLI, MRPC, could be attributed to the Contrastive Learning Objective when pre-training \fpdm\ (b).~\emrb performs considerably worse compared to \rbe on all tasks, suggesting that MLM is not robust against domain change. The possible reason behind the superior performance of $\fpdm\textit{(Cus.)}_{RoBERTa}$ is that it 
\textbf{requires only a small fraction of pre-training data} compared to what is used by domain-specific baselines such as \emrbx, hence making only \textbf{small changes in the parameter space} that helps retain open-domain knowledge while learning essential domain-specific knowledge. 
We perform an experiment to test the proposition and plot the relative change in L1-norm of different types of parameters such as attention query, key, value matrices, and dense MLP parameters (similar to \citet{noisytune}) during pre-training via MLM vs. \fpdm, as shown in Figure \ref{fig:change_params}. We observe that the relative change of parameters in \fpdm\ is about $100$ times less compared to MLM.


\subsection{Absence of document level information}

\begin{table}[H]
\small
\centering
\begin{tabular}{c|c|c|c}
\hline
\textbf{Model} &
  \textbf{F1} &
  \textbf{HA\_F1@1} &
  \textbf{HA\_F1@5} \\ \hline
\rbe       & 16.46          & 31.89             & 42.4
                                    \\
$\fpdm(Cus.)_{RoBERTa}$ (7 hier. levels) &
  \begin{tabular}[c]{@{}c@{}}17.52\\(+6.44\%)\end{tabular} &
  \begin{tabular}[c]{@{}c@{}}33.94\\(+6.43\%)\end{tabular} &
  \begin{tabular}[c]{@{}c@{}}44.96\\(+6.04\%)\end{tabular} \\ \hline

  \begin{tabular}[c]{@{}c@{}}$\fpdm(Cus.)_{RoBERTa}$ \\(w/o est. meta., tax., 7 hier. levels)\end{tabular} &
  \begin{tabular}[c]{@{}c@{}}15.39\\ (-6.5\%)\end{tabular} &
  \begin{tabular}[c]{@{}c@{}}29.83\\ (-6.46\%)\end{tabular} &
  \begin{tabular}[c]{@{}c@{}}44.35\\ (+4.6\%)\end{tabular} \\ \hline

  \begin{tabular}[c]{@{}c@{}}$\fpdm(Cus.)_{RoBERTa}$ \\(w/o est. meta., tax., 15 hier. levels)\end{tabular} &
  \textbf{\begin{tabular}[c]{@{}c@{}}18.01\\ (+9.42\%)\end{tabular}} &
  \textbf{\begin{tabular}[c]{@{}c@{}}34.89\\ (+9.41\%)\end{tabular}} &
  \textbf{\begin{tabular}[c]{@{}c@{}}47.53\\ (+12.1\%)\end{tabular}} \\
  
  \hline
\end{tabular}%

\caption{Results on TechQA Dataset in Customer Support Domain with and without established domain-specific document metadata and taxonomy}
\label{tab:Cus_wo_met_tax}
\end{table}

A pre-requisite of \fpdm\ has been the availability of document metadata and taxonomy. In this experiment, we go beyond that and derive document similarity 
via similarity based on the ROUGE-L score among documents, followed by creating a custom taxonomy of document category hierarchies using Hierarchical Topic Modeling \citep{bertopic}. Table \ref{tab:Cus_wo_met_tax} shows results on the Customer Support (see other domains' results \textbf{\emph{in Section \ref{appendix:docsup} of Appendix}}), and we can see that gives  comparable performance when considering  same number of hierarchical levels as \fpdm. However, since the taxonomy is derived using topic modeling, we are here not constrained by the number 
of hierarchies. We notice that the performance improves when a larger number of hierarchical levels are used, showing great potential for adapting \fpdm\ to any domain of interest. 
However, note that even though one can devise a (unsupervised) way to  extract triplets and document hierarchies, 
it is much more efficient to use metadata and taxonomy if and when available, as there is some time  and CPU involved in deriving content-similarity-based metrics like ROUGE-L score  due to the large size of the documents.

\subsection{Effect of using a larger backbone model for \fpdm}

\begin{table}[H]
\small
\centering
\resizebox{0.5\columnwidth}{!}{%
\begin{tabular}{c|c|c|c}
\hline
\textbf{Model} &
  \textbf{F1} &
  \textbf{HA\_F1@1} &
  \textbf{HA\_F1@5} \\ \hline

$\fpdm(Cus.)_{RoBERTa}$ &
  \begin{tabular}[c]{@{}c@{}}17.52\end{tabular} &
  \begin{tabular}[c]{@{}c@{}}33.94\end{tabular} &
  \begin{tabular}[c]{@{}c@{}}44.96\end{tabular} \\ 

  $\fpdm(Cus.)_{RL}$ &
  \textbf{\begin{tabular}[c]{@{}c@{}}18.48\\ (+5.48\%)\end{tabular}} &
  \textbf{\begin{tabular}[c]{@{}c@{}}35.8\\ (+5.48\%)\end{tabular}} &
  \textbf{\begin{tabular}[c]{@{}c@{}}47.8\\ (+6.32\%)\end{tabular}} \\
  
  \hline
\end{tabular}%
}
\caption{Results on TechQA Dataset in Customer Support Domain (RL - RoBERTa-LARGE)}
\label{tab:Cus_model_size}
\end{table}

\begin{table}[H]
	\centering
\resizebox{0.5\textwidth}{!}
{\begin{tabular}{lllcc}
\hline
\textbf{Field}       & \textbf{Task}        & \textbf{Dataset} &  \textbf{\begin{tabular}[c]{@{}c@{}}$\fpdm(Sci.)_{BERT}$\end{tabular}} & \textbf{\begin{tabular}[c]{@{}c@{}}$\fpdm(Sci.)_{BL}$\end{tabular}} \\ \hline
\multirow{4}{*}{BIO} & \multirow{3}{*}{NER} & BC5CDR         &   87.81          & \textbf{88.45 (+0.73\%)}       \\
      &     & JNLPBA &  75.84 & \textbf{76.53 (+0.91\%)} \\
      &     & NCBI-D &   84.33 & \textbf{86.18 (+2.19\%)} \\
      & REL & ChemProt  &  80.48 & \textbf{84 (+4.37\%)} \\
      \hline
CS   & REL & SciERC       &  78.95 & \textbf{80.26 (+1.66\%)} \\
\hline

Multi & CLS & SciCite      &   83.59 & \textbf{85.76 (+2.6\%)} \\
\hline
\end{tabular}
}
	\caption{Results on tasks presented in \citet{scibert} (BL - BERT-LARGE)}
	\label{tab:Sci_model_size}
\end{table}

\begin{table}[H]
\small
\centering
\begin{tabular}{c|cc}
\hline
\textbf{Model} & \multicolumn{1}{c}{\textbf{AUPR}}     &           \begin{tabular}[c]{@{}c@{}}\textbf{Precision@}\\ \textbf{80\% Recall}\end{tabular}                 \\ \hline
$\fpdm(Leg.)_{RoBERTa}$           & \begin{tabular}[c]{@{}c@{}}44.8 \end{tabular} & \begin{tabular}[c]{@{}c@{}}34.6 \end{tabular} \\ 

$\fpdm(Leg.)_{RL}$           & \textbf{\begin{tabular}[c]{@{}c@{}}45.3\\ (+1.12\%)\end{tabular}} & \textbf{\begin{tabular}[c]{@{}c@{}}39.5 \\ (+14.16\%)\end{tabular}} \\
\hline
\end{tabular}%
\caption{Results on CUAD Dataset in Legal Domain (RL - RoBERTa-LARGE)}
\label{tab:Leg_model_size}
\end{table}

Tables \ref{tab:Cus_model_size}, \ref{tab:Sci_model_size}, and \ref{tab:Leg_model_size} compare the impact of using a larger backbone compared to the one used in the proposed \fpdm\ (e.g. RoBERTa-LARGE vs. RoBERTa-BASE, BERT-LARGE vs. BERT-BASE). From the results, we can see that using a larger model as a backbone further improves results due to an increased number of trainable parameters.

\subsection{
Analysis of the interoperability of embeddings}
\label{sec:whyfpdm}
\fpdm\ shows that using input sentence embeddings during pre-training helps when using token embedding inputs during fine-tuning, as is evident from the potent downstream task performance. We analyze this  \textbf{interoperability of embeddings} by answering the following research questions (observations and experiments elaborated \emph{\textbf{in Section \ref{FT_supp} of Appendix}}) - (a). \textit{\textbf{How does}} \fpdm\ \textit{\textbf{learn local context?}} - Similar documents have very-similar local (paragraph-level) contexts, suggesting that, using document-level supervision during pre-training implicitly learns local context. Also, in an experiment, we randomly sample 500 sentences from each of the 3 domains. For each sentence, we mask a random token and calculate the change in its prediction probability on masking other tokens in the sentence. Spearman Correlation of this change between \fpdm\ and a domain-specific model pre-trained using MLM is moderately high for all domains, showing that \textbf{local context is learned by \fpdm\ to a reasonable extent}. 
(b). \textit{\textbf{Are relative representations preserved across the two embedding spaces?}} - Independent of whether inputs are sentence or token embeddings, documents are clustered in a similar manner across the two representation spaces, hence, \textbf{relative representations are preserved}. 
\section{Prior Art}
\label{sec:priorart}

\textbf{Representation Learning using self-supervised learning methods:} In recent times, downstream tasks in NLP use representation learning techniques where transformers are pre-trained on large text corpora using self-supervised learning methods like NSP \citep{bert}, MLM \citep{bert, roberta}, contrastive learning \citep{declutr, consert, cline, specter}, etc. before fine-tuning on downstream tasks. There are models pre-trained on 
domain-specific corpora such as E-Manuals \citep{nandy-etal-2021-question-answering}, legal texts \citep{legalbert}, bio-medical documents \citep{biobert}, etc.  

\textbf{Supervised Pre-training:} \citet{feng2022rethinking} proposes supervised pre-training on Leave-One-Out KNN that improves transfer to downstream tasks. CLMSM \citep{clmsm} uses recipe metadata as supervision signal for pre-training. MVP \citep{mvp} leverages labeled data from a corpus across 11 tasks for pre-training, by unifying the data into text-to-text format. The paper also states that - unsupervised pre-training likely incorporates noise that affects the downstream performance, making supervised pre-training a better alternative. CLIP \citep{clip} utilizes the pre-training task of predicting which caption goes with which image (natural language supervision), which is an efficient way to learn image representations. 

\textbf{Incorporating hierarchical information for enhancing representations:} Hierarchical information in the form of taxonomy and ontology has been used by some works to enhance learned representations. \citet{vbnet} introduces a Variational Bayes entity representation model that leverages additional hierarchical and relational information. \citet{bayesianhier} also uses a similar Bayesian approach to produce better representations, especially for rare words. 

\textbf{Intra-document Contrastive Learning:}  DeCLUTR \citep{declutr} uses a DistilRoBERTa-base \citep{distilbert} encoder. Spans overlapping or subsuming each other are considered as similar inputs, and other spans are considered as dissimilar inputs. InfoNCE Loss Function \citep{infonce1} brings representations of similar spans closer and pushes representations of dissimilar spans farther away. ConSERT \citep{consert} also uses contrastive loss, but it performs sentence augmentation using adversarial attack \citep{adversarialattack}, token shuffling, etc. It considers a sentence and its augmented counterpart to be similar, and any other sentence pair as dissimilar. CLINE \citep{cline}  
creates similar and dissimilar samples from a sentence by replacing some word(s) with their synonyms and antonyms using WordNet \citep{wordnet} and then uses contrastive loss.

\textbf{Inter-document contrastive learning:} SPECTER \citep{specter} uses a triplet margin loss to pull similar documents closer to each other, and dissimilar ones are pushed away. The document representations are obtained using a transformer encoder. However, their encoder is only able to encode a maximum of $512$ tokens of a document. 
SDR \citep{sdr} uses a self-supervised method by combining MLM loss and Contrastive Loss to learn document similarity. LinkBERT \citep{linkbert} adds a Document Relation Prediction Objective to MLM during pre-training, where the task is to predict whether two segments are contiguous, random, or from linked documents. CDLM \citep{cdlm} leverages document-level supervision by applying MLM over a set of related documents using Longformer \citep{longformer}. These works are in line with our work, but they are 
unable to tackle the important technical challenges of large input size and scalability and in turn, suffer from the problems of limited input size and high pre-training compute.

\section{Summary and Conclusion}
\label{conclusion_section}
Recent studies have repeatedly stressed the importance of domain-specific pretraining but also pointed to the costly and elaborate operation that must be undertaken to achieve reasonable performance.
This paper shows that leveraging 1) document-level semantics, and 2) interoperability of input sentence embeddings (during pre-training) and token embeddings (during fine-tuning), substantially reduces the compute requirements for domain-specific pre-training by at least 500 times,    
even while achieving better results on 6 different downstream tasks and 9 different datasets. 
The frugal pretraining technique has an important side-effect, it shows negligible {\em catastrophic forgetting} on the open-domain GLUE Benchmark.
{We also demonstrate that the existence of well-defined metadata and taxonomy is not mandatory; \fpdm\ performs effectively when discovering such metadata and taxonomy through unsupervised methods, illustrating its potential for future application across various domains.}




\subsubsection*{Limitations}

\begin{itemize}

\item \fpdm\ is robust to a wide document similarity range across several domains. However, performance in presence of high levels of noise in metadata is not guaranteed and further investigation is required to characterize it.

\item Applicability of the proposed model to decoder-only and encoder-decoder models: \fpdm\ can be extended to decoder-only models like GPT-2 \cite{gpt2}, and encoder-decoder models like BART-BASE \cite{bart}. We apply \fpdm\ using GPT-2 backbone (referred to as \fpdm$_{GPT-2}$) and the BART-BASE encoder as backbone (referred to as \fpdm$_{BART-BASE}$). Downstream task is dialogue summarization (i.e., a text generation task) on TweetSumm Dataset \cite{tweetsumm} in the Customer Support Domain. We compare it with GPT-2 and BART-BASE.

\begin{table}[H]
\small
\centering
\begin{tabular}{c|ccc}
\hline
\textbf{Model} & \textbf{ROUGE-1}     &           \textbf{ROUGE-2} & \textbf{ROUGE-L}                 \\ \hline
GPT-2           & 0.151 & 0.066 & 0.119 \\ 

           \fpdm$_{GPT-2}$ & 0.134 & 0.058 & 0.104 \\
\hline
\end{tabular}%
\caption{Results of \fpdm$_{GPT-2}$ vs. GPT-2 on TweetSumm Dataset}
\label{tab:fastdoc_gpt2}
\end{table}

\begin{table}[H]
\small
\centering
\begin{tabular}{c|ccc}
\hline
\textbf{Model} & \textbf{ROUGE-1}     &           \textbf{ROUGE-2} & \textbf{ROUGE-L}                 \\ \hline
BART-BASE           & 0.523 &	0.314	& 0.472 \\ 

           \fpdm$_{BART-BASE}$ &  0.524	& 0.315 &	0.473\\
\hline
\end{tabular}%
\caption{Results of \fpdm$_{BART-BASE}$ vs. BART-BASE on TweetSumm Dataset}
\label{tab:fastdoc_bart}
\end{table}

Tables \ref{tab:fastdoc_gpt2} and \ref{tab:fastdoc_bart} show that \fpdm\ gives poor results when using a decoder-only model as the backbone, and gives negligible improvement when using the encoder of an encoder-decoder model as the backbone. Improvement in results needs changing the architecture and document supervision objectives used in \fpdm\ to adapt to decoder and encoder-decoder models end-to-end. One way to adapt \fpdm\ to decoder model backbones is hierarchical decoding (like hierarchical encoding in \fpdm) in 2 stages - decoding special, representative sentence tokens, which are then used to decode subword tokens. This is a potential future work. 

\item Applicability of the proposed model when downstream tasks are generation tasks: Tables \ref{tab:fastdoc_gpt2} and \ref{tab:fastdoc_bart} show that \fpdm$_{GPT-2}$ and \fpdm$_{BART-BASE}$ do not perform well on a text generation task. Improvement in results could be attained in a manner mentioned above.


\end{itemize}



\subsubsection*{Broader Impact Statement}
The proposed methodology is, in general, applicable to any domain. Specifically, it can potentially be applied to user-generated text available on the web and is likely to learn patterns associated with exposure bias. This needs to be taken into consideration before applying this model to user-generated text crawled from the web. Further, like many other pre-trained language models, interpretability associated with the output is rather limited, hence users should use the output carefully.


\bibliography{main}

\begin{thebibliography}{71}
\providecommand{\natexlab}[1]{#1}
\providecommand{\url}[1]{\texttt{#1}}
\expandafter\ifx\csname urlstyle\endcsname\relax
  \providecommand{\doi}[1]{doi: #1}\else
  \providecommand{\doi}{doi: \begingroup \urlstyle{rm}\Url}\fi

\bibitem[Arumae et~al.(2020)Arumae, Sun, and Bhatia]{empmultidomain}
Kristjan Arumae, Qing Sun, and Parminder Bhatia.
\newblock An empirical investigation towards efficient multi-domain language model pre-training.
\newblock In \emph{Proceedings of the 2020 Conference on Empirical Methods in Natural Language Processing (EMNLP)}, pp.\  4854--4864, 2020.

\bibitem[Barkan et~al.(2020)Barkan, Rejwan, Caciularu, and Koenigstein]{bayesianhier}
Oren Barkan, Idan Rejwan, Avi Caciularu, and Noam Koenigstein.
\newblock {B}ayesian hierarchical words representation learning.
\newblock In \emph{Proceedings of the 58th Annual Meeting of the Association for Computational Linguistics}, pp.\  3871--3877, Online, July 2020. Association for Computational Linguistics.
\newblock \doi{10.18653/v1/2020.acl-main.356}.
\newblock URL \url{https://aclanthology.org/2020.acl-main.356}.

\bibitem[Barkan et~al.(2021)Barkan, Caciularu, Rejwan, Katz, Weill, Malkiel, and Koenigstein]{vbnet}
Oren Barkan, Avi Caciularu, Idan Rejwan, Ori Katz, Jonathan Weill, Itzik Malkiel, and Noam Koenigstein.
\newblock \emph{Representation Learning via Variational Bayesian Networks}, pp.\  78–88.
\newblock Association for Computing Machinery, New York, NY, USA, 2021.
\newblock ISBN 9781450384469.
\newblock URL \url{https://doi.org/10.1145/3459637.3482363}.

\bibitem[Beltagy et~al.(2019)Beltagy, Lo, and Cohan]{scibert}
Iz~Beltagy, Kyle Lo, and Arman Cohan.
\newblock Scibert: A pretrained language model for scientific text.
\newblock In \emph{Proceedings of the 2019 Conference on Empirical Methods in Natural Language Processing and the 9th International Joint Conference on Natural Language Processing (EMNLP-IJCNLP)}, pp.\  3615--3620, 2019.

\bibitem[Beltagy et~al.(2020)Beltagy, Peters, and Cohan]{longformer}
Iz~Beltagy, Matthew~E Peters, and Arman Cohan.
\newblock Longformer: The long-document transformer.
\newblock \emph{arXiv preprint arXiv:2004.05150}, 2020.

\bibitem[Borchert et~al.(2020)Borchert, Lohr, Modersohn, Langer, Follmann, Sachs, Hahn, and Schapranow]{metadata1}
Florian Borchert, Christina Lohr, Luise Modersohn, Thomas Langer, Markus Follmann, Jan~Philipp Sachs, Udo Hahn, and Matthieu-P Schapranow.
\newblock Ggponc: A corpus of german medical text with rich metadata based on clinical practice guidelines.
\newblock In \emph{Proceedings of the 11th International Workshop on Health Text Mining and Information Analysis}, pp.\  38--48, 2020.

\bibitem[Borchert et~al.(2022)Borchert, Lohr, Modersohn, Witt, Langer, Follmann, Gietzelt, Arnrich, Hahn, and Schapranow]{metadata2}
Florian Borchert, Christina Lohr, Luise Modersohn, Jonas Witt, Thomas Langer, Markus Follmann, Matthias Gietzelt, Bert Arnrich, Udo Hahn, and Matthieu-P Schapranow.
\newblock Ggponc 2.0-the german clinical guideline corpus for oncology: Curation workflow, annotation policy, baseline ner taggers.
\newblock In \emph{Proceedings of the Thirteenth Language Resources and Evaluation Conference}, pp.\  3650--3660, 2022.

\bibitem[Caciularu et~al.(2021)Caciularu, Cohan, Beltagy, Peters, Cattan, and Dagan]{cdlm}
Avi Caciularu, Arman Cohan, Iz~Beltagy, Matthew~E Peters, Arie Cattan, and Ido Dagan.
\newblock Cdlm: Cross-document language modeling.
\newblock In \emph{Findings of the Association for Computational Linguistics: EMNLP 2021}, pp.\  2648--2662, 2021.

\bibitem[Caruana(1993)]{hard_parameter_sharing}
Richard Caruana.
\newblock Multitask learning: A knowledge-based source of inductive bias.
\newblock In \emph{Proceedings of the Tenth International Conference on Machine Learning}, pp.\  41--48. Morgan Kaufmann, 1993.

\bibitem[Castelli et~al.(2020)Castelli, Chakravarti, Dana, Ferritto, Florian, Franz, Garg, Khandelwal, McCarley, McCawley, Nasr, Pan, Pendus, Pitrelli, Pujar, Roukos, Sakrajda, Sil, Uceda{-}Sosa, Ward, and Zhang]{techqa}
Vittorio Castelli, Rishav Chakravarti, Saswati Dana, Anthony Ferritto, Radu Florian, Martin Franz, Dinesh Garg, Dinesh Khandelwal, J.~Scott McCarley, Mike McCawley, Mohamed Nasr, Lin Pan, Cezar Pendus, John~F. Pitrelli, Saurabh Pujar, Salim Roukos, Andrzej Sakrajda, Avirup Sil, Rosario Uceda{-}Sosa, Todd Ward, and Rong Zhang.
\newblock The techqa dataset.
\newblock In Dan Jurafsky, Joyce Chai, Natalie Schluter, and Joel~R. Tetreault (eds.), \emph{Proceedings of the 58th Annual Meeting of the Association for Computational Linguistics, {ACL} 2020, Online, July 5-10, 2020}, pp.\  1269--1278. Association for Computational Linguistics, 2020.
\newblock \doi{10.18653/v1/2020.acl-main.117}.
\newblock URL \url{https://doi.org/10.18653/v1/2020.acl-main.117}.

\bibitem[Chalkidis et~al.(2019)Chalkidis, Fergadiotis, Malakasiotis, and Androutsopoulos]{eurlex}
Ilias Chalkidis, Manos Fergadiotis, Prodromos Malakasiotis, and Ion Androutsopoulos.
\newblock Large-scale multi-label text classification on {EU} legislation.
\newblock In \emph{Proceedings of the 57th Annual Meeting of the Association for Computational Linguistics}, pp.\  6314--6322, Florence, Italy, 2019. Association for Computational Linguistics.
\newblock \doi{10.18653/v1/P19-1636}.
\newblock URL \url{https://www.aclweb.org/anthology/P19-1636}.

\bibitem[Chalkidis et~al.(2020)Chalkidis, Fergadiotis, Malakasiotis, Aletras, and Androutsopoulos]{legalbert}
Ilias Chalkidis, Manos Fergadiotis, Prodromos Malakasiotis, Nikolaos Aletras, and Ion Androutsopoulos.
\newblock Legal-bert: The muppets straight out of law school.
\newblock In \emph{Findings of the Association for Computational Linguistics: EMNLP 2020}, pp.\  2898--2904, 2020.

\bibitem[Chen et~al.(2022)Chen, Yin, Shang, Jiang, Qin, Wang, Wang, Chen, Liu, and Liu]{bert2bert}
Cheng Chen, Yichun Yin, Lifeng Shang, Xin Jiang, Yujia Qin, Fengyu Wang, Zhi Wang, Xiao Chen, Zhiyuan Liu, and Qun Liu.
\newblock bert2{BERT}: Towards reusable pretrained language models.
\newblock In \emph{Proceedings of the 60th Annual Meeting of the Association for Computational Linguistics (Volume 1: Long Papers)}, pp.\  2134--2148, Dublin, Ireland, May 2022. Association for Computational Linguistics.
\newblock \doi{10.18653/v1/2022.acl-long.151}.
\newblock URL \url{https://aclanthology.org/2022.acl-long.151}.

\bibitem[Chen et~al.(2020)Chen, Kornblith, Norouzi, and Hinton]{ntxent}
Ting Chen, Simon Kornblith, Mohammad Norouzi, and Geoffrey Hinton.
\newblock A simple framework for contrastive learning of visual representations.
\newblock In \emph{International conference on machine learning}, pp.\  1597--1607. PMLR, 2020.

\bibitem[Clark et~al.(2019)Clark, Celikyilmaz, and Smith]{swms}
Elizabeth Clark, Asli Celikyilmaz, and Noah~A Smith.
\newblock Sentence mover’s similarity: Automatic evaluation for multi-sentence texts.
\newblock In \emph{Proceedings of the 57th Annual Meeting of the Association for Computational Linguistics}, pp.\  2748--2760, 2019.

\bibitem[Cohan et~al.(2019)Cohan, Ammar, van Zuylen, and Cady]{scicite}
Arman Cohan, Waleed Ammar, Madeleine van Zuylen, and Field Cady.
\newblock Structural scaffolds for citation intent classification in scientific publications.
\newblock In \emph{Proceedings of the 2019 Conference of the North American Chapter of the Association for Computational Linguistics: Human Language Technologies, Volume 1 (Long and Short Papers)}, pp.\  3586--3596, 2019.

\bibitem[Cohan et~al.(2020)Cohan, Feldman, Beltagy, Downey, and Weld]{specter}
Arman Cohan, Sergey Feldman, Iz~Beltagy, Doug Downey, and Daniel~S. Weld.
\newblock {SPECTER: Document-level Representation Learning using Citation-informed Transformers}.
\newblock In \emph{ACL}, 2020.

\bibitem[Devlin et~al.(2019)Devlin, Chang, Lee, and Toutanova]{bert}
Jacob Devlin, Ming-Wei Chang, Kenton Lee, and Kristina Toutanova.
\newblock {BERT}: Pre-training of deep bidirectional transformers for language understanding.
\newblock In \emph{Proceedings of the 2019 Conference of the North {A}merican Chapter of the Association for Computational Linguistics: Human Language Technologies, Volume 1 (Long and Short Papers)}, pp.\  4171--4186, Minneapolis, Minnesota, June 2019. Association for Computational Linguistics.
\newblock \doi{10.18653/v1/N19-1423}.
\newblock URL \url{https://aclanthology.org/N19-1423}.

\bibitem[Do{\u{g}}an et~al.(2014)Do{\u{g}}an, Leaman, and Lu]{ncbi}
Rezarta~Islamaj Do{\u{g}}an, Robert Leaman, and Zhiyong Lu.
\newblock Ncbi disease corpus: a resource for disease name recognition and concept normalization.
\newblock \emph{Journal of biomedical informatics}, 47:\penalty0 1--10, 2014.

\bibitem[Feigenblat et~al.(2021)Feigenblat, Gunasekara, Sznajder, Joshi, Konopnicki, and Aharonov]{tweetsumm}
Guy Feigenblat, Chulaka Gunasekara, Benjamin Sznajder, Sachindra Joshi, David Konopnicki, and Ranit Aharonov.
\newblock {TWEETSUMM} - a dialog summarization dataset for customer service.
\newblock In Marie-Francine Moens, Xuanjing Huang, Lucia Specia, and Scott Wen-tau Yih (eds.), \emph{Findings of the Association for Computational Linguistics: EMNLP 2021}, pp.\  245--260, Punta Cana, Dominican Republic, November 2021. Association for Computational Linguistics.
\newblock \doi{10.18653/v1/2021.findings-emnlp.24}.
\newblock URL \url{https://aclanthology.org/2021.findings-emnlp.24}.

\bibitem[Feng et~al.(2022)Feng, Jiang, Tang, Jin, and Gao]{feng2022rethinking}
Yutong Feng, Jianwen Jiang, Mingqian Tang, Rong Jin, and Yue Gao.
\newblock Rethinking supervised pre-training for better downstream transferring.
\newblock In \emph{International Conference on Learning Representations}, 2022.
\newblock URL \url{https://openreview.net/forum?id=Jjcv9MTqhcq}.

\bibitem[Geng et~al.(2021)Geng, Lebret, and Aberer]{legalroberta}
Saibo Geng, R{\'e}mi Lebret, and Karl Aberer.
\newblock Legal transformer models may not always help.
\newblock \emph{arXiv preprint arXiv:2109.06862}, 2021.

\bibitem[Ginzburg et~al.(2021)Ginzburg, Malkiel, Barkan, Caciularu, and Koenigstein]{sdr}
Dvir Ginzburg, Itzik Malkiel, Oren Barkan, Avi Caciularu, and Noam Koenigstein.
\newblock Self-supervised document similarity ranking via contextualized language models and hierarchical inference.
\newblock In \emph{Findings of the Association for Computational Linguistics: ACL-IJCNLP 2021}, pp.\  3088--3098, Online, August 2021. Association for Computational Linguistics.
\newblock \doi{10.18653/v1/2021.findings-acl.272}.
\newblock URL \url{https://aclanthology.org/2021.findings-acl.272}.

\bibitem[Giorgi et~al.(2021)Giorgi, Nitski, Wang, and Bader]{declutr}
John~M. Giorgi, Osvald Nitski, Bo~Wang, and Gary~D. Bader.
\newblock Declutr: Deep contrastive learning for unsupervised textual representations.
\newblock In Chengqing Zong, Fei Xia, Wenjie Li, and Roberto Navigli (eds.), \emph{Proceedings of the 59th Annual Meeting of the Association for Computational Linguistics and the 11th International Joint Conference on Natural Language Processing, {ACL/IJCNLP} 2021, (Volume 1: Long Papers), Virtual Event, August 1-6, 2021}, pp.\  879--895. Association for Computational Linguistics, 2021.
\newblock \doi{10.18653/v1/2021.acl-long.72}.
\newblock URL \url{https://doi.org/10.18653/v1/2021.acl-long.72}.

\bibitem[Grootendorst(2022)]{bertopic}
Maarten Grootendorst.
\newblock Bertopic: Neural topic modeling with a class-based tf-idf procedure.
\newblock \emph{arXiv preprint arXiv:2203.05794}, 2022.

\bibitem[Gururangan et~al.(2020)Gururangan, Marasovi{\'c}, Swayamdipta, Lo, Beltagy, Downey, and Smith]{dontstop}
Suchin Gururangan, Ana Marasovi{\'c}, Swabha Swayamdipta, Kyle Lo, Iz~Beltagy, Doug Downey, and Noah~A. Smith.
\newblock Don{'}t stop pretraining: Adapt language models to domains and tasks.
\newblock In \emph{Proceedings of the 58th Annual Meeting of the Association for Computational Linguistics}, pp.\  8342--8360, Online, July 2020. Association for Computational Linguistics.
\newblock \doi{10.18653/v1/2020.acl-main.740}.
\newblock URL \url{https://aclanthology.org/2020.acl-main.740}.

\bibitem[Hendrycks et~al.(2021)Hendrycks, Burns, Chen, and Ball]{cuad}
Dan Hendrycks, Collin Burns, Anya Chen, and Spencer Ball.
\newblock Cuad: An expert-annotated nlp dataset for legal contract review.
\newblock \emph{NeurIPS}, 2021.

\bibitem[Hu et~al.(2022)Hu, yelong shen, Wallis, Allen-Zhu, Li, Wang, Wang, and Chen]{lora}
Edward~J Hu, yelong shen, Phillip Wallis, Zeyuan Allen-Zhu, Yuanzhi Li, Shean Wang, Lu~Wang, and Weizhu Chen.
\newblock Lo{RA}: Low-rank adaptation of large language models.
\newblock In \emph{International Conference on Learning Representations}, 2022.
\newblock URL \url{https://openreview.net/forum?id=nZeVKeeFYf9}.

\bibitem[Huang et~al.(2023)Huang, Wang, and Yang]{finbert}
Allen~H Huang, Hui Wang, and Yi~Yang.
\newblock Finbert: A large language model for extracting information from financial text.
\newblock \emph{Contemporary Accounting Research}, 40\penalty0 (2):\penalty0 806--841, 2023.

\bibitem[Karamanolakis et~al.(2020)Karamanolakis, Ma, and Dong]{taxonomy2}
Giannis Karamanolakis, Jun Ma, and Xin~Luna Dong.
\newblock Txtract: Taxonomy-aware knowledge extraction for thousands of product categories.
\newblock In \emph{Proceedings of the 58th Annual Meeting of the Association for Computational Linguistics}, pp.\  8489--8502, 2020.

\bibitem[Kim et~al.(2004)Kim, Ohta, Tsuruoka, Tateisi, and Collier]{jnlpba}
Jin-Dong Kim, Tomoko Ohta, Yoshimasa Tsuruoka, Yuka Tateisi, and Nigel Collier.
\newblock Introduction to the bio-entity recognition task at jnlpba.
\newblock In \emph{Proceedings of the international joint workshop on natural language processing in biomedicine and its applications}, pp.\  70--75. Citeseer, 2004.

\bibitem[Kringelum et~al.(2016)Kringelum, Kjaerulff, Brunak, Lund, Oprea, and Taboureau]{chemprot}
Jens Kringelum, Sonny~Kim Kjaerulff, S{\o}ren Brunak, Ole Lund, Tudor~I Oprea, and Olivier Taboureau.
\newblock Chemprot-3.0: a global chemical biology diseases mapping.
\newblock \emph{Database: The Journal of Biological Databases and Curation}, 2016, 2016.

\bibitem[Kurakin et~al.(2016)Kurakin, Goodfellow, Bengio, et~al.]{adversarialattack}
Alexey Kurakin, Ian Goodfellow, Samy Bengio, et~al.
\newblock Adversarial examples in the physical world, 2016.

\bibitem[Lan et~al.(2019)Lan, Chen, Goodman, Gimpel, Sharma, and Soricut]{albert}
Zhenzhong Lan, Mingda Chen, Sebastian Goodman, Kevin Gimpel, Piyush Sharma, and Radu Soricut.
\newblock Albert: A lite bert for self-supervised learning of language representations.
\newblock In \emph{International Conference on Learning Representations}, 2019.

\bibitem[Le \& Mikolov(2014)Le and Mikolov]{doc2vec}
Quoc Le and Tomas Mikolov.
\newblock Distributed representations of sentences and documents.
\newblock In \emph{International conference on machine learning}, pp.\  1188--1196. PMLR, 2014.

\bibitem[Lee et~al.(2020)Lee, Yoon, Kim, Kim, Kim, So, and Kang]{biobert}
Jinhyuk Lee, Wonjin Yoon, Sungdong Kim, Donghyeon Kim, Sunkyu Kim, Chan~Ho So, and Jaewoo Kang.
\newblock Biobert: a pre-trained biomedical language representation model for biomedical text mining.
\newblock \emph{Bioinformatics}, 36\penalty0 (4):\penalty0 1234--1240, 2020.

\bibitem[Lewis et~al.(2020)Lewis, Liu, Goyal, Ghazvininejad, Mohamed, Levy, Stoyanov, and Zettlemoyer]{bart}
Mike Lewis, Yinhan Liu, Naman Goyal, Marjan Ghazvininejad, Abdelrahman Mohamed, Omer Levy, Veselin Stoyanov, and Luke Zettlemoyer.
\newblock {BART}: Denoising sequence-to-sequence pre-training for natural language generation, translation, and comprehension.
\newblock In Dan Jurafsky, Joyce Chai, Natalie Schluter, and Joel Tetreault (eds.), \emph{Proceedings of the 58th Annual Meeting of the Association for Computational Linguistics}, pp.\  7871--7880, Online, July 2020. Association for Computational Linguistics.
\newblock \doi{10.18653/v1/2020.acl-main.703}.
\newblock URL \url{https://aclanthology.org/2020.acl-main.703}.

\bibitem[Li et~al.(2016)Li, Sun, Johnson, Sciaky, Wei, Leaman, Davis, Mattingly, Wiegers, and Lu]{bc5cdr}
Jiao Li, Yueping Sun, Robin~J. Johnson, Daniela Sciaky, Chih-Hsuan Wei, Robert Leaman, Allan~Peter Davis, Carolyn~J. Mattingly, Thomas~C. Wiegers, and Zhiyong Lu.
\newblock {BioCreative V CDR task corpus: a resource for chemical disease relation extraction}.
\newblock \emph{Database}, 2016, 05 2016.
\newblock ISSN 1758-0463.
\newblock \doi{10.1093/database/baw068}.
\newblock URL \url{https://doi.org/10.1093/database/baw068}.
\newblock baw068.

\bibitem[Lin(2004)]{rouge}
Chin-Yew Lin.
\newblock {ROUGE}: A package for automatic evaluation of summaries.
\newblock In \emph{Text Summarization Branches Out}, pp.\  74--81, Barcelona, Spain, July 2004. Association for Computational Linguistics.
\newblock URL \url{https://aclanthology.org/W04-1013}.

\bibitem[Lipscomb(2000)]{metadata3}
Carolyn~E Lipscomb.
\newblock Medical subject headings (mesh).
\newblock \emph{Bulletin of the Medical Library Association}, 88\penalty0 (3):\penalty0 265, 2000.

\bibitem[Liu et~al.(2019)Liu, Ott, Goyal, Du, Joshi, Chen, Levy, Lewis, Zettlemoyer, and Stoyanov]{roberta}
Yinhan Liu, Myle Ott, Naman Goyal, Jingfei Du, Mandar Joshi, Danqi Chen, Omer Levy, Mike Lewis, Luke Zettlemoyer, and Veselin Stoyanov.
\newblock Roberta: A robustly optimized bert pretraining approach.
\newblock \emph{arXiv preprint arXiv:1907.11692}, 2019.

\bibitem[Loshchilov \& Hutter(2018)Loshchilov and Hutter]{adamw}
Ilya Loshchilov and Frank Hutter.
\newblock Decoupled weight decay regularization.
\newblock In \emph{International Conference on Learning Representations}, 2018.

\bibitem[Loza~Mencia et~al.(2010)Loza~Mencia, Fürnkranz, and loza]{eurlex_darmstadt}
Eneldo Loza~Mencia, Johannes Fürnkranz, and loza.
\newblock Eur-lex dataset, 2010.
\newblock URL \url{https://tudatalib.ulb.tu-darmstadt.de/handle/tudatalib/2937}.

\bibitem[Luan et~al.(2018)Luan, He, Ostendorf, and Hajishirzi]{scierc}
Yi~Luan, Luheng He, Mari Ostendorf, and Hannaneh Hajishirzi.
\newblock Multi-task identification of entities, relations, and coreference for scientific knowledge graph construction.
\newblock In \emph{Proceedings of the 2018 Conference on Empirical Methods in Natural Language Processing}, pp.\  3219--3232, 2018.

\bibitem[Malkiel et~al.(2022)Malkiel, Ginzburg, Barkan, Caciularu, Weill, and Koenigstein]{metricbert}
Itzik Malkiel, Dvir Ginzburg, Oren Barkan, Avi Caciularu, Yoni Weill, and Noam Koenigstein.
\newblock Metricbert: Text representation learning via self-supervised triplet training.
\newblock In \emph{ICASSP 2022-2022 IEEE International Conference on Acoustics, Speech and Signal Processing (ICASSP)}, pp.\  1--5. IEEE, 2022.

\bibitem[Margiotta et~al.(2022)Margiotta, Croce, Rotoloni, Cacciamani, and Basili]{taxonomy1}
Daniele Margiotta, Danilo Croce, Marco Rotoloni, Barbara Cacciamani, and Roberto Basili.
\newblock Knowledge-based neural pre-training for intelligent document management.
\newblock In Stefania Bandini, Francesca Gasparini, Viviana Mascardi, Matteo Palmonari, and Giuseppe Vizzari (eds.), \emph{AIxIA 2021 -- Advances in Artificial Intelligence}, pp.\  564--579, Cham, 2022. Springer International Publishing.
\newblock ISBN 978-3-031-08421-8.

\bibitem[Mikolov et~al.(2013)Mikolov, Chen, Corrado, and Dean]{word2vec}
Tom{\'{a}}s Mikolov, Kai Chen, Greg Corrado, and Jeffrey Dean.
\newblock Efficient estimation of word representations in vector space.
\newblock In Yoshua Bengio and Yann LeCun (eds.), \emph{1st International Conference on Learning Representations, {ICLR} 2013, Scottsdale, Arizona, USA, May 2-4, 2013, Workshop Track Proceedings}, 2013.
\newblock URL \url{http://arxiv.org/abs/1301.3781}.

\bibitem[Miller(1995)]{wordnet}
George~A. Miller.
\newblock Wordnet: A lexical database for english.
\newblock \emph{Commun. ACM}, 38\penalty0 (11):\penalty0 39–41, nov 1995.
\newblock ISSN 0001-0782.
\newblock \doi{10.1145/219717.219748}.
\newblock URL \url{https://doi.org/10.1145/219717.219748}.

\bibitem[Nandy et~al.(2021)Nandy, Sharma, Maddhashiya, Sachdeva, Goyal, and Ganguly]{nandy-etal-2021-question-answering}
Abhilash Nandy, Soumya Sharma, Shubham Maddhashiya, Kapil Sachdeva, Pawan Goyal, and NIloy Ganguly.
\newblock Question answering over electronic devices: A new benchmark dataset and a multi-task learning based {QA} framework.
\newblock In \emph{Findings of the Association for Computational Linguistics: EMNLP 2021}, pp.\  4600--4609, Punta Cana, Dominican Republic, November 2021. Association for Computational Linguistics.
\newblock URL \url{https://aclanthology.org/2021.findings-emnlp.392}.

\bibitem[Nandy et~al.(2023)Nandy, Kapadnis, Goyal, and Ganguly]{clmsm}
Abhilash Nandy, Manav Kapadnis, Pawan Goyal, and Niloy Ganguly.
\newblock {CLMSM}: A multi-task learning framework for pre-training on procedural text.
\newblock In Houda Bouamor, Juan Pino, and Kalika Bali (eds.), \emph{Findings of the Association for Computational Linguistics: EMNLP 2023}, pp.\  8793--8806, Singapore, December 2023. Association for Computational Linguistics.
\newblock \doi{10.18653/v1/2023.findings-emnlp.589}.
\newblock URL \url{https://aclanthology.org/2023.findings-emnlp.589}.

\bibitem[Nogueira et~al.(2019)Nogueira, Lin, and Epistemic]{bm25}
Rodrigo Nogueira, Jimmy Lin, and AI~Epistemic.
\newblock From doc2query to doctttttquery.
\newblock \emph{Online preprint}, 2019.

\bibitem[Oh~Song et~al.(2016)Oh~Song, Xiang, Jegelka, and Savarese]{margin1}
Hyun Oh~Song, Yu~Xiang, Stefanie Jegelka, and Silvio Savarese.
\newblock Deep metric learning via lifted structured feature embedding.
\newblock In \emph{Proceedings of the IEEE conference on computer vision and pattern recognition}, pp.\  4004--4012, 2016.

\bibitem[Ostendorff et~al.(2022)Ostendorff, Rethmeier, Augenstein, Gipp, and Rehm]{impr_specter}
Malte Ostendorff, Nils Rethmeier, Isabelle Augenstein, Bela Gipp, and Georg Rehm.
\newblock Neighborhood contrastive learning for scientific document representations with citation embeddings.
\newblock \emph{arXiv preprint arXiv:2202.06671}, 2022.

\bibitem[Radford et~al.(2019)Radford, Wu, Child, Luan, Amodei, and Sutskever]{gpt2}
Alec Radford, Jeff Wu, Rewon Child, David Luan, Dario Amodei, and Ilya Sutskever.
\newblock Language models are unsupervised multitask learners.
\newblock \emph{OpenAI Blog}, 2019.

\bibitem[Radford et~al.(2021)Radford, Kim, Hallacy, Ramesh, Goh, Agarwal, Sastry, Askell, Mishkin, Clark, et~al.]{clip}
Alec Radford, Jong~Wook Kim, Chris Hallacy, Aditya Ramesh, Gabriel Goh, Sandhini Agarwal, Girish Sastry, Amanda Askell, Pamela Mishkin, Jack Clark, et~al.
\newblock Learning transferable visual models from natural language supervision.
\newblock In \emph{International conference on machine learning}, pp.\  8748--8763. PMLR, 2021.

\bibitem[Rajpurkar et~al.(2016)Rajpurkar, Zhang, Lopyrev, and Liang]{squadv1}
Pranav Rajpurkar, Jian Zhang, Konstantin Lopyrev, and Percy Liang.
\newblock Squad: 100,000+ questions for machine comprehension of text.
\newblock In \emph{Proceedings of the 2016 Conference on Empirical Methods in Natural Language Processing}, pp.\  2383--2392, 2016.

\bibitem[Rajpurkar et~al.(2018)Rajpurkar, Jia, and Liang]{squadv2}
Pranav Rajpurkar, Robin Jia, and Percy Liang.
\newblock Know what you don’t know: Unanswerable questions for squad.
\newblock In \emph{Proceedings of the 56th Annual Meeting of the Association for Computational Linguistics (Volume 2: Short Papers)}, pp.\  784--789, 2018.

\bibitem[Reimers \& Gurevych(2019)Reimers and Gurevych]{sbert}
Nils Reimers and Iryna Gurevych.
\newblock Sentence-bert: Sentence embeddings using siamese bert-networks.
\newblock In \emph{Proceedings of the 2019 Conference on Empirical Methods in Natural Language Processing and the 9th International Joint Conference on Natural Language Processing (EMNLP-IJCNLP)}, pp.\  3982--3992, 2019.

\bibitem[Sanh et~al.(2019)Sanh, Debut, Chaumond, and Wolf]{distilbert}
Victor Sanh, Lysandre Debut, Julien Chaumond, and Thomas Wolf.
\newblock Distilbert, a distilled version of bert: smaller, faster, cheaper and lighter.
\newblock \emph{arXiv preprint arXiv:1910.01108}, 2019.

\bibitem[Silla \& Freitas(2011)Silla and Freitas]{hierclassfn}
Carlos~N Silla and Alex~A Freitas.
\newblock A survey of hierarchical classification across different application domains.
\newblock \emph{Data Mining and Knowledge Discovery}, 22\penalty0 (1):\penalty0 31--72, 2011.

\bibitem[Sohn(2016)]{infonce1}
Kihyuk Sohn.
\newblock Improved deep metric learning with multi-class n-pair loss objective.
\newblock \emph{Advances in neural information processing systems}, 29:\penalty0 1857--1865, 2016.

\bibitem[Tang et~al.(2023)Tang, Li, Zhao, and Wen]{mvp}
Tianyi Tang, Junyi Li, Wayne~Xin Zhao, and Ji-Rong Wen.
\newblock {MVP}: Multi-task supervised pre-training for natural language generation.
\newblock In Anna Rogers, Jordan Boyd-Graber, and Naoaki Okazaki (eds.), \emph{Findings of the Association for Computational Linguistics: ACL 2023}, pp.\  8758--8794, Toronto, Canada, July 2023. Association for Computational Linguistics.
\newblock \doi{10.18653/v1/2023.findings-acl.558}.
\newblock URL \url{https://aclanthology.org/2023.findings-acl.558}.

\bibitem[Tenney et~al.(2019)Tenney, Das, and Pavlick]{prob_bert}
Ian Tenney, Dipanjan Das, and Ellie Pavlick.
\newblock {BERT} rediscovers the classical {NLP} pipeline.
\newblock In \emph{Proceedings of the 57th Annual Meeting of the Association for Computational Linguistics}, pp.\  4593--4601, Florence, Italy, July 2019. Association for Computational Linguistics.
\newblock \doi{10.18653/v1/P19-1452}.
\newblock URL \url{https://aclanthology.org/P19-1452}.

\bibitem[van Aken et~al.(2019)van Aken, Winter, L\"{o}ser, and Gers]{probe_qa}
Betty van Aken, Benjamin Winter, Alexander L\"{o}ser, and Felix~A. Gers.
\newblock How does bert answer questions? a layer-wise analysis of transformer representations.
\newblock In \emph{Proceedings of the 28th ACM International Conference on Information and Knowledge Management}, CIKM '19, pp.\  1823–1832, New York, NY, USA, 2019. Association for Computing Machinery.
\newblock ISBN 9781450369763.
\newblock \doi{10.1145/3357384.3358028}.
\newblock URL \url{https://doi.org/10.1145/3357384.3358028}.

\bibitem[Wang et~al.(2018)Wang, Singh, Michael, Hill, Levy, and Bowman]{glue}
Alex Wang, Amanpreet Singh, Julian Michael, Felix Hill, Omer Levy, and Samuel Bowman.
\newblock {GLUE}: A multi-task benchmark and analysis platform for natural language understanding.
\newblock In \emph{Proceedings of the 2018 {EMNLP} Workshop {B}lackbox{NLP}: Analyzing and Interpreting Neural Networks for {NLP}}, pp.\  353--355, Brussels, Belgium, November 2018. Association for Computational Linguistics.
\newblock \doi{10.18653/v1/W18-5446}.
\newblock URL \url{https://aclanthology.org/W18-5446}.

\bibitem[Wang et~al.(2021)Wang, Ding, Li, and Zheng]{cline}
Dong Wang, Ning Ding, Piji Li, and Haitao Zheng.
\newblock {CLINE}: Contrastive learning with semantic negative examples for natural language understanding.
\newblock In \emph{Proceedings of the 59th Annual Meeting of the Association for Computational Linguistics and the 11th International Joint Conference on Natural Language Processing (Volume 1: Long Papers)}, pp.\  2332--2342, Online, August 2021. Association for Computational Linguistics.
\newblock \doi{10.18653/v1/2021.acl-long.181}.
\newblock URL \url{https://aclanthology.org/2021.acl-long.181}.

\bibitem[Weinberger \& Saul(2009)Weinberger and Saul]{margin2}
Kilian~Q Weinberger and Lawrence~K Saul.
\newblock Distance metric learning for large margin nearest neighbor classification.
\newblock \emph{Journal of machine learning research}, 10\penalty0 (2), 2009.

\bibitem[Wu et~al.(2022)Wu, Wu, Qi, and Huang]{noisytune}
Chuhan Wu, Fangzhao Wu, Tao Qi, and Yongfeng Huang.
\newblock {N}oisy{T}une: A little noise can help you finetune pretrained language models better.
\newblock In \emph{Proceedings of the 60th Annual Meeting of the Association for Computational Linguistics (Volume 2: Short Papers)}, pp.\  680--685, Dublin, Ireland, May 2022. Association for Computational Linguistics.
\newblock \doi{10.18653/v1/2022.acl-short.76}.
\newblock URL \url{https://aclanthology.org/2022.acl-short.76}.

\bibitem[Yan et~al.(2021)Yan, Li, Wang, Zhang, Wu, and Xu]{consert}
Yuanmeng Yan, Rumei Li, Sirui Wang, Fuzheng Zhang, Wei Wu, and Weiran Xu.
\newblock Consert: {A} contrastive framework for self-supervised sentence representation transfer.
\newblock In Chengqing Zong, Fei Xia, Wenjie Li, and Roberto Navigli (eds.), \emph{Proceedings of the 59th Annual Meeting of the Association for Computational Linguistics and the 11th International Joint Conference on Natural Language Processing, {ACL/IJCNLP} 2021, (Volume 1: Long Papers), Virtual Event, August 1-6, 2021}, pp.\  5065--5075. Association for Computational Linguistics, 2021.
\newblock \doi{10.18653/v1/2021.acl-long.393}.
\newblock URL \url{https://doi.org/10.18653/v1/2021.acl-long.393}.

\bibitem[Yasunaga et~al.(2022)Yasunaga, Leskovec, and Liang]{linkbert}
Michihiro Yasunaga, Jure Leskovec, and Percy Liang.
\newblock Linkbert: Pretraining language models with document links.
\newblock In \emph{Proceedings of the 60th Annual Meeting of the Association for Computational Linguistics (Volume 1: Long Papers)}, pp.\  8003--8016, 2022.

\bibitem[Zhang et~al.(2019)Zhang, Wei, and Zhou]{hibert}
Xingxing Zhang, Furu Wei, and Ming Zhou.
\newblock {HIBERT}: Document level pre-training of hierarchical bidirectional transformers for document summarization.
\newblock In \emph{Proceedings of the 57th Annual Meeting of the Association for Computational Linguistics}, pp.\  5059--5069, Florence, Italy, July 2019. Association for Computational Linguistics.
\newblock \doi{10.18653/v1/P19-1499}.
\newblock URL \url{https://aclanthology.org/P19-1499}.

\end{thebibliography}
\bibliographystyle{tmlr}

\appendix

\section*{Appendix}

The Appendix is organized in the same sectional format as the main paper. The additional material of
a section is put in the corresponding section of the Appendix so that it becomes easier for the reader
to find the relevant information. Some sections and subsections may not have supplementary material so only their name is mentioned. The page numbers are in continuation from the Main Paper's end page number.


\section{Introduction}

\section{\fpdm\ Framework}
\label{pre_train_suppl}

\begin{figure}[!thb]
    \centering
    \includegraphics[width=0.7\textwidth]{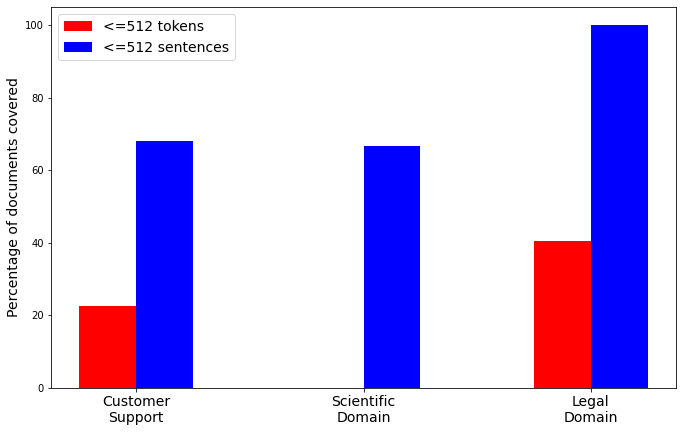}
    \caption{Percentage of documents encoded entirely by RoBERTa-BASE encoder when the input is $512$ tokens vs. $512$ sentences (The \textcolor{red}{red} bar of ``Scientific Domain" has a negligible height, and is hence, not visible.)}
    \label{fig:appendix_token_distro}
\end{figure}

\begin{figure}[!thb]
    \centering
    \includegraphics[width=0.64\textwidth]{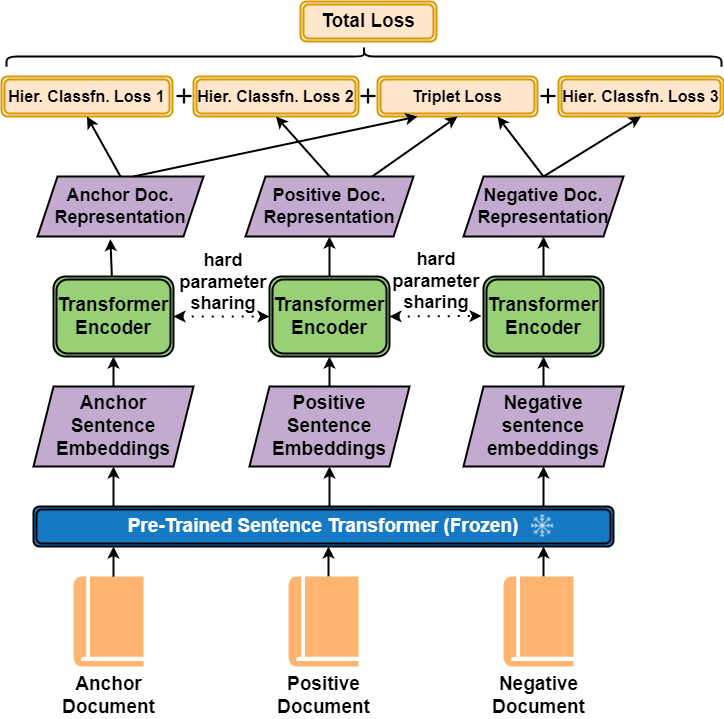}
    \caption{Depiction of \fpdm. Anchor, Positive, and Negative Documents are encoded using a Sentence Transformer, followed by a transformer encoder, to give document representations. A combination of Triplet and Hierarchical Classification Losses is used to get the Total Loss}
    \label{fig:arch}
\end{figure}



Figure \ref{fig:appendix_token_distro} shows the percentage of documents encoded entirely by RoBERTa-BASE encoder when the input is $512$ tokens vs. $512$ sentences. 

Figure \ref{fig:arch} shows a detailed overview of the \fpdm~Framework.

\section{Pre-training Setup}

\subsection{Pre-training in the Customer Support Domain}
\label{pre_train_cust_supp}
\begin{table}[!htb]
\centering
\small

\begin{tabular}{p{0.1\textwidth}p{0.4\textwidth}}
\hline
\textbf{Product Category} & \textbf{Hierarchical categories assigned to the product}                                                                                                                                                                  \\ \hline
blu-ray player            & Electronics \textgreater~ Audio \textgreater~ Audio Accessories \textgreater~ MP3 Player Accessories \textgreater~ MP3 Player \& Mobile Phone Accessory Sets \\ \hline
stereo equalizer        & \textbf{Electronics} \textgreater~ \textbf{Audio} \textgreater~ Audio Players \& Recorders \textgreater~ Stereo Systems                                                                  \\ \hline
laptop docking station        & \textbf{Electronics} \textgreater~ Electronics Accessories \textgreater~ Computer Accessories \textgreater~ Laptop Docking Stations                                                                  \\ \hline
hot beverage maker        & Home \& Garden \textgreater~ Kitchen \& Dining \textgreater~ Kitchen Appliance Accessories \textgreater~ Coffee Maker \& Espresso Machine Accessories \textgreater~ Coffee Maker \& Espresso Machine Replacement Parts                                                                  \\ \hline
\end{tabular}%

\caption{Examples of product categories and their corresponding hierarchical categories assigned with the help of the Google Product Taxonomy. Similar products have similar hierarchies}
\label{tab:taxonomy_mapping_examples}
\end{table}

Table \ref{tab:taxonomy_mapping_examples} shows $4$ examples of E-manual product categories and the hierarchies  assigned to them with the help of the GPrT. We can see that more similar products tend to have more similar hierarchies.

\noindent\textbf{Mapping E-Manual product category to hierarchy: }It may so happen that the product category of the E-Manual does not have an exact match with any leaf category in the GPrT. In that case, we map it to that leaf category,
where  
 cosine similarity of mean word embeddings~\citep{word2vec} of the product category and the hierarchy's last two entities is the highest. This choice gives qualitatively good mapping when measured using human evaluation.

 \noindent\textbf{Triplet Count in Customer Support} Note that we do not double the number of triplets by swapping anchor and positive in case of Customer Support, as doing so leads to very similar results, while taking double the compute. This is because in case of Customer Support, E-Manuals are written based on the product category metadata, while the metadata for Scientific and Legal Domains are assigned after the document is written. Thus, the product category in Customer Support is much more precise and well-defined compared to primary category and overlap in EUROVOC Concepts used in Scientific and Legal Domains respectively, thus requiring lesser data.

\section{Downstream Datasets/Tasks}
\label{appendix:downstream}

\textbf{Datasets used in our work}

Table \ref{tab:appendix_datasets} lists all the downstream datasets used in our work, along with their corresponding tasks and domains. 

\begin{table}[!thb]
\small
\centering
\begin{tabular}{c|c|c}
\hline
\textbf{Domain} & \textbf{Task}                                                 & \textbf{Dataset} \\ \hline
\multirow{2}{*}{\begin{tabular}[c]{@{}c@{}}Customer\\ Support\end{tabular}} &
  \begin{tabular}[c]{@{}c@{}}single-span\\ QA\end{tabular} &
  TechQA \\ \cline{2-3} 
 &
  \begin{tabular}[c]{@{}c@{}}multi-span\\ QA\end{tabular} &
  \begin{tabular}[c]{@{}c@{}}S10 QA\\ Dataset\end{tabular} \\ \hline
\multirow{6}{*}{\begin{tabular}[c]{@{}c@{}}Scientific\\ Domain\end{tabular}} &
  \multirow{3}{*}{NER} &
  BC5CDR \\
                &                                                               & JNLPBA           \\
                &                                                               & NCBI-D           \\ \cline{2-3} 
 &
  \multirow{2}{*}{\begin{tabular}[c]{@{}c@{}}Relation\\ Classification\end{tabular}} &
  ChemProt \\
                &                                                               & SciERC           \\ \cline{2-3} 
                & \begin{tabular}[c]{@{}c@{}}Text\\ Classification\end{tabular} & SciCite          \\ \hline
\begin{tabular}[c]{@{}c@{}}Legal\\ Domain\end{tabular} &
  \begin{tabular}[c]{@{}c@{}}Contract \\ Review \\ (Span-based\\ Clause\\ Extraction)\end{tabular} &
  CUAD \\ \hline
\end{tabular}%
\caption{List of all the datasets along with their corresponding tasks and domains.}
\label{tab:appendix_datasets}
\end{table}

\subsection{Customer Support}
\label{appendix:s10_desc}

\subsubsubsection{\bf S10 Question Answering Dataset.} The S10 QA Dataset~\citep{nandy-etal-2021-question-answering} consists of $904$ question-answer pairs curated from the Samsung S10 Smartphone E-Manual\footnote{\url{https://bit.ly/36bqs5E}}, along with additional information on the section of the E-Manual containing the answer. 
However, the answer might not be a continuous span, i.e., the answer may be present in the form of non-contiguous sentences of a section. The tasks of section and answer retrieval are performed. The dataset is divided in the ratio of 7:2:1 into training, validation, and test sets, respectively.

\subsection{Scientific Domain}

\subsection{Legal Domain}
\label{appendix:legal_downstream}

The dataset is split 80/20 into train/test, with a small validation set for the preliminary experiments to perform hyperparameter grid search. 

\section{Experiments and Results}
\label{appendix:exps}

Here we report certain extra experiments which could not be accommodated in the main paper due to want of space. We also report \textbf{additional ablation analysis} on the TechQA Dataset below.


\subsection{Customer Support Domain}
\label{cust_supp_exps_supp}

\noindent \textbf{Baselines: }\ul{Details on SPECTER} - When pre-training on E-Manuals, instead of initializing the encoder with SciBERT \citep{scibert} (as in \citet{specter}),   we initialize  the model using \embert \citep{nandy-etal-2021-question-answering}, sample about the same number of E-Manual triplets stated in \citet{specter} as inputs (using product category information), and use the first $512$ tokens per input E-Manual.

\subsection*{Fine-tuning Setup}

\noindent \textbf{Fine-tuning on SQuAD 2.0 \citep{squadv2}.} SQuAD 2.0 is a span-based open-domain reading comprehension dataset, consisting of $130,319$ training, $11,873$ dev, and $8,862$ test QA pairs. Before fine-tuning on the task-specific dataset, we fine-tune the encoder on the SQuAD 2.0 training set, as it has been shown to improve the performance on QA tasks \citep{techqa}. 
The hyperparameters used are the same as mentioned in \citet{squadv2}.

\noindent \textbf{Fine-tuning on TechQA Dataset: }The encoder 
is fine-tuned on the TechQA Dataset with the same training architecture used when fine-tuning on SQuAD 2.0. Since this is a QA task, a question and one of the candidate technotes separated by a special token is the input. If the technote contains the answer, the target output is the start and end token of the answer,  and it is unanswerable otherwise. The hyperparameters used are the ones mentioned in the default implementation\footnote{\url{https://github.com/IBM/techqa} - Apache-2.0 License} of~\citet{techqa}. 

\textbf{Fine-tuning on S10 QA Dataset: }The S10 Dataset is accompanied by $2$ sub-tasks - (a) \textbf{Section Retrieval} - given the question and top $10$ candidate sections retrieved using BM25 IR Method \citep{bm25}\footnote{BM25 is better than TF-IDF used in \citet{nandy-etal-2021-question-answering}}, the task is to find out the section that contains the answer. (b) \textbf{Answer Retrieval} - Given a question and the relevant E-Manual section, the task is to retrieve the answer to the question. 
For section retrieval, a (question, candidate section) pair separated by special tokens (`[CLS]' and `[SEP]' in case of BERT \citep{bert} and `\textless s\textgreater' and `\textless /s\textgreater' in case of RoBERTa \citep{roberta}) is input to the model, 
and the ground truth is $0/1$ depending on whether the section contains the answer or not. Similarly, in the case of answer retrieval, the question paired with a sentence of the section containing the answer (separated by special tokens) is the input to the model, 
and the ground truth is $0/1$ depending on whether the sentence is a part of the answer or not. In both sub-tasks, the classification token's (`[CLS]' or `\textless s\textgreater') encoder output is fed to a linear layer (followed by Softmax function) to get a probability value. 
Fine-tuning on each of the sub-tasks yields separate models which are used during inference time for completion of the respective task. 

For all the fine-tuning experiments on S10 QA Dataset, we use a batch size of $16$ (except for the pre-trained DeCLUTR model with $\text{DistilRoBERTa}_{\text{BASE}}$ backbone, where a batch size of 32 is used), and train for $4$ epochs with an AdamW optimizer \citep{adamw} and an initial learning rate of $4 \times 10^{-5}$, that decays linearly.

\begin{table}[t]
\small
\centering
\begin{tabular}{l|ccc}
\hline
                                                                                                                                & \textbf{F1}    & \textbf{HA\_F1@1} & \textbf{HA\_F1@5} \\ \hline
 \bert       & \color{red}8.63          & \color{red}16.72             & \color{red}22.52                         \\
                                                                                                        \rbe       & \color{red}13.98          & \color{red}27.1             & \color[HTML]{008000}43.02
                        \\            Longformer & \color{red}15.39 & \color{red}29.82 & \color{red}42                                                                    \\ \embert          & \color{red}10.1  & \color{red}19.56    & \color{red}29.87                       \\
                                                                                                        \emrb       & \color{red}13.62 & \color{red}26.38    & \color{red}38.67              \\
                                                                                                        DeCLUTR                 & \color{red}12.52 & \color{red}24.26    & \color{red}29.59                       \\
                                                                                                        ConSERT                 & \color{red}10.78 & \color{red}20.88    & \color[HTML]{008000}31.55                       \\
                                                                                                        SPECTER                 & \color{red}0.69  & \color{red}1.34     & \color{red}7.24                       \\ \hline
 \fpdm$(Cus.)_{BERT}$(hier.)           & \color{red}9.12  & \color{red}17.68    & \color{red}26.52                    \\
                                                                                                        \fpdm$(Cus.)_{BERT}$(triplet)                 & \color{red}10.76          & \color{red}20.84             & \color{red}31.81                       \\
                                                                                                        \fpdm$(Cus.)_{BERT}$ & \color{red}7.8  & \color{red}15.11    & \color{red}24.78                       \\ \hline 
 \fpdm$(Cus.)_{RoBERTa}$(hier.)           & \color{red}13.83 &        \color{red}26.8        & \color{red}37.84                    \\
                                                                                                        \fpdm$(Cus.)_{RoBERTa}$(triplet)                 & \color{red}12.93 &        \color{red}25.06        & \color{red}40.84             \\
                                                                                                        \fpdm$(Cus.)_{RoBERTa}$ & \color{red}14.89 &        \color{red}28.85 &        \color{red}39.04                       \\ \hline 
\end{tabular}%

\caption{Results for the QA downstream task on the TechQA Dataset, \textbf{without intermediate SQuAD 2.0 fine-tuning} (Values in \textcolor{red}{red}/\textcolor[HTML]{008000}{green} indicate if the values are \textcolor{red}{less than}/\textcolor[HTML]{008000}{greater than} the values got using intermediate SQuAD 2.0 fine-tuning) 
}
\label{tab:techqa_results_no_squad}
\end{table}

\subsection*{Results}
\noindent \textbf{Performance on TechQA Dataset}

\noindent \textbf{Analyzing impact of fine-tuning on SQuAD 2.0: }Table \ref{tab:techqa_results_no_squad} shows the results on the TechQA Dataset without intermediate fine-tuning on SQuAD 2.0 Dataset. Intermediate SQuAD 2.0 fine-tuning definitively improves results for $6$ out of $8$ baselines, and all the \fpdm~variants. 

\noindent \textbf{Additional Ablation Analysis}

We perform the following ablations on $\fpdm(Cus.)_{RoBERTa}$ - (1) We pre-train both the lower and higher-level encoders and fine-tune the lower encoder. This is referred to as $\fpdm(Cus.)_{RoBERTa}(FULL)$ (2) We replace the lower encoder sRoBERTa (sentence transformer) with RoBERTa-BASE (still keeping its weights frozen) and refer to it as $\fpdm(Cus.)_{RoBERTa} (lower-RoBERTa)$.

\begin{table}[!htb]
\centering
\small
\begin{tabular}{l|ccc}
\hline
                                                                                                                                & \textbf{F1}    & \textbf{HA\_F1@1} & \textbf{HA\_F1@5} \\ \hline
                                                                                                        
                                                                                                        \fpdm$(Cus.)_{RoBERTa}$ (FULL) & 17.4	& 33.71	& \textbf{46.23}                      \\ 
                                                                                                        
                                                                                                        \fpdm$(Cus.)_{RoBERTa}$ (lower-RoBERTa) & 15.76 &	30.54	& 42.52                      \\ 
                                                                                                        \fpdm$(Cus.)_{RoBERTa}$ & \textbf{17.52} & \textbf{33.94}    & 44.96                      \\ \hline                                                                                                        
                                                                                                        
\end{tabular}%
\caption{Additional Ablation Analysis on TechQA Dataset.} 
\label{tab:techqa_additional_ablation}
\end{table}

Table \ref{tab:techqa_additional_ablation} shows the results corresponding to the ablations and \fpdm\ on TechQA Dataset. We can see that \fpdm\ performs better than $\fpdm(Cus.)_{RoBERTa} (lower-RoBERTa)$ on all 3 metrics, and gives better F1 and HA\_F1@1 than $\fpdm(Cus.)_{RoBERTa}(FULL)$.

\begin{figure}[t]
    \centering
    \includegraphics[width=0.64\textwidth]{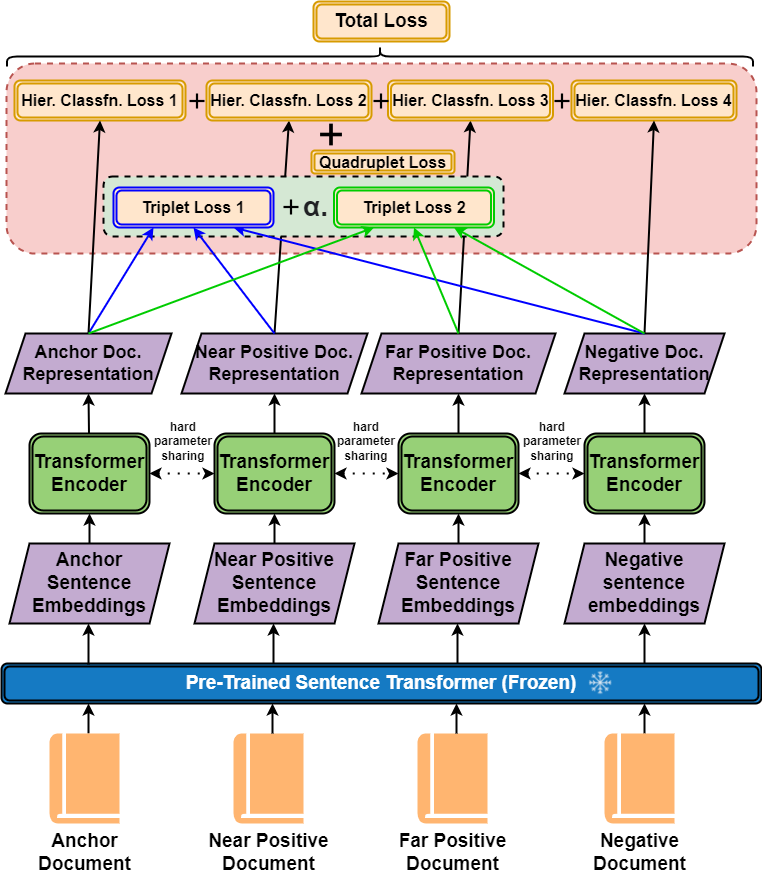}
    \caption{\fpdm$(Q)$ Pre-training Architecture - It is similar to that of \fpdm. Differences are - (1) Instead of using a Triplet Loss Function (as in \fpdm), a Quadruplet Loss Function is used in this case. (2) Anchor, Near Positive, Far Positive, and Negative Documents are taken as inputs.}
    \label{fig:fpdm-q}
\end{figure}

\noindent \textbf{An alternative to the Triplet Loss: }We also observe the effect of using an alternative for triplet loss. We use ``quadruplet loss'', which breaks similarity into 3 categories instead of the binary notion in triplet loss. We sample 4 documents per input - Anchor(A), near positive(NP), far positive(FP), negative(N). NP is most similar to anchor, followed by FP, and N. E.g. in Customer Support, NP has same brand, category as that of anchor, FP has same category, different brand, and N has neither same brand nor category. Quadruplet Loss, denoted by $\mathcal{L}_{t}$, can be mathematically stated as
        \begin{equation}
            \text{ $\mathcal{L}_{q} = \mathcal{L}_{t}(A, NP, N) + K. \mathcal{L}_{t}(A, FP, N)$}
            \label{eq:docQuadSim}
        \end{equation}

, where $K (0<K<1)$ is a constant to reduce weight of the second loss term in Equation \ref{eq:docQuadSim}, as distance between A and NP is to be reduced more than that between A and FP. We denote this variation of \fpdm\ as \fpdm$(Q)$, as depicted in Figure \ref{fig:fpdm-q}. From Table \ref{tab:quadruplet_comparison}, we observe that \fpdm\ performs better than \fpdm$(Q)$ for $K = 0.1, 0.5$ w.r.t all 3 metrics on the TechQA Dataset.

\begin{table}[H]
\small
\centering
\begin{tabular}{l|ccc}
\hline
 & \textbf{F1} & \textbf{HA\_F1@1} & \textbf{HA\_F1@5} \\
\hline
\fpdm$(Cus.)_{RoBERTa}$(Q, K=0.1) & 16.1 & 31.2 & 43.04 \\
\fpdm$(Cus.)_{RoBERTa}$(Q, K=0.5) & 15.72 & 30.45 & 40.42 \\
\fpdm$(Cus.)_{RoBERTa}$ & \textbf{17.52} & \textbf{33.94} & \textbf{44.96} \\
\hline
\end{tabular}%
\caption{Performance of \fpdm$(Q)$ vs. \fpdm\ on the TechQA Dataset}
\label{tab:quadruplet_comparison}
\end{table}

\subsubsection*{\bf Performance on the S10 QA Dataset}



Table~\ref{tab:s10_downstream} shows the performance of baselines and proposed variants on Section Retrieval and Answer Retrieval tasks on S10 QA Dataset. Results are reported on the test set, similar to \citet{nandy-etal-2021-question-answering}.
For \textbf{Section Retrieval} we report HITS@K - the percentage of questions for which, the section containing the ground truth answer is one of the top $K$ retrieved sections. We report values for $K = 1, 3$.
In \textbf{Answer Retrieval} a single answer is retrieved, hence  HA\_F1@1 is reported. 
The other metrics reported are 
(a). ROUGE-L \footnote{used \url{https://pypi.org/project/py-rouge/}} \citep{rouge}, 
and (b). Sentence and Word Mover Similarity (S+WMS) \footnote{used \url{https://github.com/eaclark07/sms}} \citep{swms}\footnote{ROUGE-L F1, S+WMS are reported, as all questions in the S10 QA Dataset are answerable, and these metrics make sense when each question has a ground truth answer. 
}. 
    
    

We draw the following inferences from Table~\ref{tab:s10_downstream} -
(1) Longformer  does not perform well, as global attention does not help in learning local contexts required for answer retrieval.  (2) Among the baselines, \emrb gives the best HITS@1, \rbe gives the best ROUGE-L F1, and DeCLUTR gives the best S+WMS score and HA\_F1@1. This shows that MLM and span-based contrastive learning help extract non-contiguous answer spans.  (3) Similar to TechQA, $\fpdm\textit{(Cus.)}_{RoBERTa}$ variants perform better than $\fpdm\textit{(Cus.)}_{BERT}$ variants. (4) Compared to TechQA, HA\_F1@1 scores on S10 QA Dataset are better, as answering questions from a single device is easier than answering questions from diverse sources. (5) In Section Retrieval, $\fpdm\textit{(Cus.)}_{RoBERTa}$ gives the best HITS@1, suggesting that, pre-training using document-level supervision helps generalize to a device not seen during pre-training. (6) \ul{$\fpdm\textit{(Cus.)}_{RoBERTa}$ gives the best ROUGE-L F1 and the second-best S+WMS score}, while \ul{$\fpdm\textit{(Cus.)}_{RoBERTa}(hier.)$ gives the best S+WMS score, and the second-best ROUGE-L F1 and HA\_F1@1}, as mapping documents to hierarchical labels during pre-training can generalize to the context of a device E-Manual not seen during pre-training. The hierarchy is particularly robust due to the variety in hierarchical labels.
Combining it with triplet loss improves the lexical context, as can be seen from the value of ROUGE-L F1. 

\begin{table}[!htb]
\centering
\small
\begin{tabular}{l|cc|ccc}
\hline
   &
  \multicolumn{2}{c|}{\textbf{HITS@K}} &
  \multicolumn{3}{c}{\textbf{Answer Retrieval}} \\ \cline{2-6} 
 
   &
  $K=1$ &
  $K=3$ &
  \textbf{\begin{tabular}[c]{@{}c@{}}ROUGE\\-L F1\end{tabular}} &
  \textbf{\begin{tabular}[c]{@{}c@{}}S+\\WMS\end{tabular}} &
  \textbf{\begin{tabular}[c]{@{}c@{}}HA\_\\F1@1\end{tabular}} \\ \hline

  \bert &
  76.67 &
  91.11 &
  0.792 &
  0.411 &
  44.87 \\
\rbe &
  80 &
  \textbf{93.33} &
  0.812 &
  0.454 &
  45.39 \\
  Longformer & 75.56 &\textbf{93.33} & 0.768 & 0.415 & 41.1
  \\
\embert &
  \ul{81.11} &
  \textbf{93.33} &
  0.8 &
  0.429 &
  44.25 \\
\emrb &
  \textbf{82.22} &
  \textbf{93.33} &
  \ul{0.82} &
  0.444 &
  44.73 \\
  DeCLUTr &
  76.67 &
  \ul{92.22} &
  0.818 &
  0.455 &
  \textbf{46.71} 
  \\
  ConSERT &
  78.89 &
  \ul{92.22} &
  0.778 &
  0.389 &
  40.85 \\
  SPECTER &
  77.78 &
  \textbf{93.33} &
  0.802 &
  0.429 &
  43.59 \\
  \hline
  \fpdm$(Cus.)_{BERT}$(hier.) &
  \ul{81.11} &
  \textbf{93.33} &
  0.791 &
  0.427 &
  43.18 \\
  \fpdm$(Cus.)_{BERT}$(triplet) &
  77.78 &
  \textbf{93.33} &
  0.798 &
  0.419 &
  42.93 \\
  \fpdm$(Cus.)_{BERT}$ &
  78.89 &
  \textbf{93.33} &
  0.79 &
  0.412 &
  41.75 \\ \hline
  \fpdm$(Cus.)_{RoBERTa}$(hier.) &
  78.89 &
  \ul{92.22} &
  \ul{0.82} &
  \textbf{0.478} &
  \ul{46.69} \\
  
  \fpdm$(Cus.)_{RoBERTa}$(triplet)&
  80 &
  \textbf{93.33} &
  0.811 &
  0.437 &
  43.78 \\
  \fpdm$(Cus.)_{RoBERTa}$ &
  \textbf{82.22} &
  \textbf{93.33} &
  \textbf{0.828} &
  \ul{0.463} &
  46.22 \\ \hline
\end{tabular}%
\caption{Results on the S10 QA Dataset (Best value for each metric is marked in \textbf{bold}, while the second-best value is underlined). 
}
\label{tab:s10_downstream}
\end{table}

\if{0}
\textbf{Qualitative Analysis of answers predicted by a proposed variant and a baseline}

We discuss qualitative results with $2$ questions from TechQA Dataset and $1$ question from S10 QA Dataset and the answers  $\fpdm\textit{(Cus.)}_{RoBERTa}$ and a consistently well-performing baseline \emrb\footnote{Although Longformer performs well on TechQA, it does not perform well on S10 QA Dataset. For compactness, we chose a well-performing baseline. However, illustrations will be similar.} provide for each question. These questions, ground truth, and predicted answers are listed in Table \ref{tab:qual_analysis_full} in \emph{\textbf{Section \ref{cust_supp_exps_supp} of Appendix}}. 
 The first question is a procedural question (`How' type), where both models give extra information w.r.t the ground truth. However, $\fpdm\textit{(Cus.)}_{RoBERTa}$  performs better in extracting the exact number corresponding to the `Fix' which \emrb misses. 
The second question is in essence two questions together where one is  procedural (`How type') and the other is  factual (`Is' type) question. Both the models output short answers that have minimal overlap with the ground truth, suggesting that it is difficult to answer multiple questions of different types at one go. However, $\fpdm\textit{(Cus.)}_{RoBERTa}$ is nearer to the answer, as it suggests the issue is  related to `WS-Proxies' (mentioned in the ground truth), but that does not appear in the baseline's answer.  The third question is a location-based question (`Where' type). $\fpdm\textit{(Cus.)}_{RoBERTa}$ answers it almost correctly, with just some extra information on the paragraph containing the answer, whereas the baseline confuses `fingerprint list' for `fingerprint recognition'.
 The observations point to the proposition that 
 $\fpdm_{RoBERTa}$ performs better 
 at extracting numerical entities, tackling multiple questions in a sample, and answering location-based questions. 

 \fi

\subsubsection*{\bf Qualitative Analysis of answers predicted by a proposed variant and a baseline}

We discuss qualitative results with $2$ questions from TechQA Dataset and $1$ question from S10 QA Dataset and the answers  $\fpdm_{RoBERTa}$ and a consistently well-performing baseline \emrb\footnote{Note that even though Longformer performs well on TechQA, it does not perform well on S10 QA Dataset. For compactness, we chose only one consistently well performing baseline. However, the illustrations will be similar.} provide for each question. These questions, ground truth and predicted answers are listed in Table \ref{tab:qual_analysis_full}. 
 The first question is a procedural question (`How' type), where both models give extra information w.r.t the ground truth. However, $\fpdm_{RoBERTa}$  performs better in extracting the exact number corresponding to the `Fix' which \emrb misses. 
The second question is in essence two questions together where one is  procedural (`How type') and the other is  factual (`Is' type) question. Both the models output short answers that have minimal overlap with the ground truth, suggesting that it is difficult to answer multiple questions of different types at one go. However, $\fpdm_{RoBERTa}$ is nearer to the answer, as it suggests the issue is  related to `WS-Proxies' (mentioned in the ground truth), but that does not appear in the baseline's answer.  The third question is a location-based question (`Where' type). $\fpdm_{RoBERTa}$ answers it almost correctly, with just some extra information on the paragraph containing the answer, whereas the baseline confuses `fingerprint list' for `fingerprint recognition'.
 The observations point to the proposition that 
 \textbf{$\fpdm_{RoBERTa}$ performs better} 
 \textbf{at extracting numerical entities, tackling multiple questions in a sample, and answering location-based questions.}

\begin{table*}[!thb]
\centering
{%
\scriptsize
\begin{tabular}{p{0.06\textwidth}p{0.21\textwidth}p{0.2\textwidth}p{0.2\textwidth}p{0.2\textwidth}}
\hline
\textbf{Dataset} & \textbf{Question} & \textbf{Ground Truth Answer} & \fpdm$(Cus.)_{RoBERTa}$ & \emrb \\ \hline
                                 & We are using Portal Server 8.5 and when starting the Portal server we see following error:\textbackslash{}n\textbackslash{}n0000005e MBeanStarter E  com.ibm.wps.scripting.server. MBeanStarter loadExtensions failed to load extension: key: ServerExt.applications / value: com.ibm.wps.scripting.server. ApplicationServerExtensions How can we resolve this error?  
                                 & A code fix for this issue is integrated into the WebSphere\textbackslash{}n   Portal \& WCM Combined Cumulative Fix 05 (PI31198 {[}http://www-01.ibm.com/support/ docview.wss? 
 uid=swg1PI31198{]}) for Portal 8.5.0.0 available from Fix Central:
                                 & \color[HTML]{008000}{A code fix for this issue is integrated into} \color[HTML]{008000}{the WebSphere\textbackslash{}n   Portal \& WCM Combined}\color[HTML]{008000}{Cumulative Fix 05 (PI31198 {[}http://www-01} \color[HTML]{008000}{.ibm.com/support/docview. wss?uid=swg1PI31198{]})} \color[HTML]{008000}{for Portal\textbackslash{}n   8.5.0.0 available from Fix} \color[HTML]{008000}{Central:} \textcolor{red}{$<$LONG URL$>$}
                                 & \color[HTML]{008000}{A code fix for this issue is integrated into} \color[HTML]{008000}{the WebSphere\textbackslash{}n   Portal \& WCM Combined} \color[HTML]{008000}{Cumulative Fix \textcolor{red}{06} (\textcolor{red}{PI36060} {[}http://www-01} \color[HTML]{008000}{.ibm.com/support/docview. wss?uid=swg1\textcolor{red}{PI36060}{]})} \color[HTML]{008000}{for Portal\textbackslash{}n   8.5.0.0 available from Fix}\color[HTML]{008000}{ Central:} 
                    \textcolor{red}{$<$LONG URL$>$}
                                 \\
                                 
TechQA & I have made changes to a WSDL stored in DP local filesystem. How can I make the WSP reflect this change since DataPower Gateway 7.1.0 WebService Proxy WebGUI doesn't offer the \textbackslash{}"Refresh\textbackslash{}" button anymore? Is there a best practice to refresh the WSP if I modified my WSDL?                                                            & In a \textbf{WS Proxy}, you can refresh WSDL from an URL using WSDL Cache policy. However, if the WSDL is hosted on the DataPower appliance, such as in local:///, rather than on an external URL, the WSDL Cache Policy does not apply. For instance, changes made to local:///test.wsdl are not picked up by \textbf{WS Proxy} .... To refresh a WSDL in the local:/// directory, disable and re-enable the service. & {\color[HTML]{FE0000} To minimize the impact of this issue, consider creating additional \textbf{WS-Proxi}es to reduce the number of WSDLs per proxy.}                                                                                                                                                                                                                                                                                                                                                                                                                                                                        & {\color[HTML]{FE0000} Add/remove the WSDL from the WSP Gateway each time you change the WSDL. 2) Add one char to the user summary and then backspace it. The apply button will be enabled now. 3) Add a WSDL Cache Policy.}\\                                                                                                                                                                                                                                                                                                                                                                                       \hline
S10 QA Dataset                   & I need the registered fingerprint list. Where can I find this?                                                                                                                                                                                                                                                                                                                      & From Settings, tap Biometrics and security \textgreater Fingerprints for the following options: The list of registered fingerprints is at the top of this list. You can tap a fingerprint to remove or rename it.                                                                                                                                                                                                                                                                                                                                                                                                                                                                                                                                                              & \textcolor{red}{Add, delete, and rename fingerprints.} \color[HTML]{008000}{From} \color[HTML]{008000}{Settings, tap Biometrics and security \textgreater} \color[HTML]{008000}{Fingerprints for the following options: The} \color[HTML]{008000}{list of registered fingerprints is at the top of} \color[HTML]{008000}{this list. You can tap} \color[HTML]{008000}{a fingerprint to remove or rename it.}                                                                                                                                                                                                                                                                                                                                                          &\textcolor{red}{Use fingerprint recognition to verify your identity in supported apps and actions. From Settings, tap Biometrics and security \textgreater Fingerprints. This is not available on the Galaxy S10e. This is only available on the Galaxy S10e.}                                                                                                                                           
\\ \hline

\end{tabular}%
}
\caption{Qualitative Analysis of answers predicted by $\fpdm\textit{(Cus.)}_{RoBERTa}$ and the \emrb baseline on the TechQA and S10 QA Datasets. (The parts in {\color[HTML]{008000}green} overlap with the ground truth, and the parts in {\color[HTML]{FF0000}red} do not overlap.)}
\label{tab:qual_analysis_full}
\end{table*}

In an effort to understand the reason behind \fpdm\ being better at extracting numerical entities or answering location-based questions, we see how well are such entities shared by anchor and positive documents compared to anchor and negative documents in Table \ref{tab:appendix_qual_ents}.

\begin{table}[H]
\small
\centering
\begin{tabular}{c|ccc}
\hline
 &
  \textbf{\begin{tabular}[c]{@{}c@{}}Number of common entities\\between anchor and positive\end{tabular}} &
  \textbf{\begin{tabular}[c]{@{}c@{}}Number of common entities\\between anchor and negative\end{tabular}} & \textbf{\begin{tabular}[c]{@{}c@{}}Number of common entities\\in (anchor, positive) and\\not in negative\end{tabular}} \\ \hline
Numerical &
 20.7  & 10.2 & 13.3
   \\
Noun Phrase & 61.5
    & 6.6 & 57.9
   \\ \hline
\end{tabular}%
\caption{Analysis of the distribution of numerical and noun phrase entities across document triplets}
\label{tab:appendix_qual_ents}
\end{table}

We observe that overlap between anchor and positive is more than that between anchor and negative. Also, number of entities in (positive, anchor) and not in negative is more than the number of common entities between anchor and negative. Hence, numerical entities and locations (subset of noun phrases) are shared across highly similar documents, which is captured in the contrastive learning task performed during pre-training \fpdm. Hence, \fpdm\ is proficient at answering numerical entity and location-based questions.


\subsection{Scientific Domain}
\label{appendix:scientific_exps}

Since recent works have used citations as a similarity signal, we report the performance of \fpdm\ using citations as a similarity signal in Table \ref{tab:appendix_scibert_paper_tasks}. This gives a satisfactory performance, showing that \fpdm\ works on different metadata types. However, on average, a system using citations does not perform as well as when using ``primary category". In line with the recent works on Contrastive Learning, we apply \fpdm\ on triplets sampled using citations as a similarity signal and denote it as $\fpdm\textit{(Sci.-Cit.)}_{BERT}$.
We can see in Table \ref{tab:appendix_scibert_paper_tasks} that on an average, $\fpdm\textit{(Sci.)}_{BERT}$ performs better than $\fpdm\textit{(Sci.-Cit.)}_{BERT}$.

\begin{table}[H]
\small
\centering

\begin{tabular}{lllccc}
\hline
\textbf{Field}       & \textbf{Task}        & \textbf{Dataset} & $\fpdm\textit{(Sci.-Cit.)}_{BERT}$  & $\fpdm\textit{(Sci.)}_{BERT}$ \\ \hline
\multirow{4}{*}{BIO} & \multirow{3}{*}{NER} & BC5CDR           &   87.55  & 87.81                 \\
      &     & JNLPBA & 75.9 & 75.84 \\
      &     & NCBI-D & 85.12 &  84.33 \\
      & REL & ChemProt  & 73.8 & 80.48 \\
      \hline
CS   & REL & SciERC       & 80.8 & 78.95 \\
\hline
Multi & CLS & SciCite      & 84.13 &  83.59\\
\hline
 &  &  AVERAGE & 81.22 & \textbf{81.83} \\ \hline
\end{tabular}
\caption{$\fpdm\textit{(Sci.)}_{BERT}$ vs. $\fpdm\textit{(Sci.-Cit.)}_{BERT}$ in tasks presented in \citet{scibert}. We report macro F1 for NER (span-level), and for REL and CLS (sentence-level), except for ChemProt, where we report micro F1.}
\label{tab:appendix_scibert_paper_tasks}
\end{table}

\noindent \textbf{Comparison with GPT-3.5}

We compare \fpdm\ with GPT-3.5 in the zero and one-shot settings for some tasks in the Scientific Domain in Tables \ref{tab:Sci_chatgpt_zero} and \ref{tab:Sci_chatgpt_one} respectively. We can see that our proposed \fpdm\ performs much better compared to the highly capable and much larger GPT-3.5 in both zero and one-shot settings.

\begin{table}[H]
\small
	\centering
\begin{tabular}{lllcc}
\hline
\textbf{Field}       & \textbf{Task}        & \textbf{Dataset} &  \textbf{\begin{tabular}[c]{@{}c@{}}$\fpdm(Sci.)_{BERT}$\end{tabular}} & \textbf{\begin{tabular}[c]{@{}c@{}}GPT-3.5 (Zero-Shot)\end{tabular}} \\ \hline
\multirow{4}{*}{BIO} & \multirow{3}{*}{NER} & BC5CDR         &   \textbf{87.81}          & 56.04 (-36.18\%)       \\
      &     & JNLPBA &  \textbf{75.84} & 41.25 (-45.61\%) \\
      &     & NCBI-D &   \textbf{84.33} &  50.49 (-40.13\%) \\

      & REL & ChemProt  &  \textbf{80.48} & 34.16 (-57.56\%) \\
      \hline


\end{tabular}
	\caption{Results of \fpdm\ vs. zero-shot GPT-3.5  on some of the tasks presented in \citet{scibert}.}
	\label{tab:Sci_chatgpt_zero}
\end{table}

\begin{table}[H]
\small
	\centering
\begin{tabular}{lllcc}
\hline
\textbf{Field}       & \textbf{Task}        & \textbf{Dataset} &  \textbf{\begin{tabular}[c]{@{}c@{}}$\fpdm(Sci.)_{BERT}$\end{tabular}} & \textbf{\begin{tabular}[c]{@{}c@{}}GPT-3.5 (One-Shot)\end{tabular}} \\ \hline

BIO      & REL & ChemProt  &  \textbf{80.48} & 48.64 (-39.56\%) \\
      \hline


\end{tabular}
	\caption{Results of \fpdm\ vs. one-shot GPT-3.5  on some of the tasks presented in \citet{scibert}.}
	\label{tab:Sci_chatgpt_one}
\end{table}

\subsection{Legal Domain}
\label{appendix:legal_exps}

\section{Pre-training Compute of \fpdm~relative to the baselines}
\label{appendix:pretr_compute}

\textbf{Choice of GPU-Hours as a metric for measuring pre-training compute}

\fpdm\ and the baselines in Table \ref{tab:compute} use a BERT/RoBERTa backbone, making modules of sharding, data parallelism, etc. uniform across models. Hence, GPU-Hours is appropriate for compute. Also, several works on pre-training such as \citet{bert, roberta, specter, declutr} report the number of GPUs and the time needed for pre-training, i.e., they support the metric of GPU-Hours.

Also, we compare the metrics of GPU-Hours and FLOPS (Floating-point operations per second) in Table \ref{tab:appendix_flops}. Note that FLOPS is another reliable metric for measuring computer performance\footnote{\url{https://kb.iu.edu/d/apeq}}, and hence, training compute. 

\begin{table}[H]
\small
\centering
\begin{tabular}{c|c|c}
\hline
 &
  \textbf{FLOPS} &
  \textbf{GPU-Hours} \\ \hline
\emrb &
 $4.46 \times 10^18$  & 980
   \\
$\fpdm(Cus.)_{RoBERTa}$ & $2.56 \times 10^15$
    & 0.75
   \\ \hline
Speedup & 1745
    & 1307
   \\ \hline
\end{tabular}%
\caption{Comparison of the speedup obtained using \fpdm\ in GPU-Hours vs. FLOPS}
\label{tab:appendix_flops}
\end{table}

We observe that the compute speedup (or reduction) in \fpdm\ compared to the \emrb\ baseline is very close for the two metrics of GPU-Hours and FLOPS, suggesting that GPU-Hours, like FLOPS, is indeed a reliable metric for measuring pre-training compute.

\textbf{Clarification of calculation of pre-training compute of \scibert}

We would like to present the following evidence accompanied by suitable reasoning in support of the calculated value of the pre-training compute -

\begin{enumerate}
    \item The blog referred to in Footnote 5 of \citet{scibert}  (\url{https://timdettmers.com/2018/10/17/tpus-vs-gpus-for-transformers-bert/}), titled ``TPUs vs GPUs for Transformers (BERT)", discusses the compute requirements for BERT-LARGE and BERT-BASE using different GPU and TPU configurations and specifications. A line from a section of the blog titled ``BERT Training Time Estimate for GPUs” states - ``On an 8 GPU machine for V100/RTX 2080 Tis with any software and any parallelization algorithm (PyTorch, TensorFlow) one can expect to train BERT-LARGE in 21 days or 34 days". This does not match the sentence in the footnote, which suggests that it is expected to take 40-70 days for pre-training on an 8 GPU machine. Hence, we believe that the sentence in the footnote does not correspond to BERT-LARGE, rather, it intuitively corresponds to \scibert. The blog was referred to give the reader an idea of how a comparison of GPU and TPU is made.
    \item We must mention here that the TPU versions used in \citet{scibert} and \citet{bert} are in all probability different. \citet{scibert} reports its pre-training time corresponding to TPU v3, whereas \citet{bert} does not mention the exact version of the Cloud TPU used. Also, according to the Google Cloud TPU Release Notes (\url{https://cloud.google.com/tpu/docs/release-notes#October_10_2018}) we see that the TPU v3 was first introduced (in beta release) on October 10, 2018. However, the first version of the BERT Paper was added to ArXiv (\url{https://arxiv.org/abs/1810.04805v1}) on October 11, 2018, just 1 day after the beta release of TPU v3 - indicating that, some earlier version of TPU was used for the pretraining experiments.
\end{enumerate}

\section{Analysis and Ablation}
\subsection{Catastrophic Forgetting in open-domain}
\label{appendix_cf}

\textbf{Hyperparameters: }For all such experiments, we fine-tune for $10$ epochs, with a learning rate of $3\times10^{-5}$, input sequence length of $512$, and batch size of $32$. For a task, the best development set results across all epochs is reported.



\subsection{Parameter-Efficient training}
\label{appendix:peft}

The trainable parameters of the encoder during fine-tuning are the same as that in the domain-specific pre-training stage. Hence, as an ablation, we explore the impact of a reduced number of trainable parameters during pre-training by incorporating Parameter-Efficient Training.

\begin{table}[!thb]
\small
	\centering
\begin{tabular}{lllcc}
\hline
\textbf{Field}       & \textbf{Task}        & \textbf{Dataset} &  \textbf{\begin{tabular}[c]{@{}c@{}}$\fpdm(Sci.)_{BERT}$\end{tabular}} & \textbf{\begin{tabular}[c]{@{}c@{}}$\fpdm(Sci.)_{BERT}(LoRA)$\end{tabular}} \\ \hline
\multirow{4}{*}{BIO} & \multirow{3}{*}{NER} & BC5CDR         &   87.81          & \textbf{88.59 (+0.89\%)}       \\
      &     & JNLPBA &  \textbf{75.84} & 61.24 (-19.25\%) \\
      &     & NCBI-D &   \textbf{84.33} & 39.57 (-53.08\%) \\

      & REL & ChemProt  &  \textbf{80.48} & 74.32 (-7.65\%) \\
      \hline
CS   & REL & SciERC       &  \textbf{78.95} & 62.19 (-21.23\%) \\
\hline

Multi & CLS & SciCite      &   \textbf{83.59} & 81.35 (-2.68\%) \\
\hline      

\end{tabular}
	\caption{Results of \fpdm\ trained using LoRA vs. proposed \fpdm\ on tasks presented in \citet{scibert}.}
	\label{tab:Sci_param_eff}
\end{table}

Table \ref{tab:Sci_param_eff} shows the results of the parameter-efficient training technique of LoRA (Low-Rank Adaptation) \citep{lora} to observe the effect of using a reduced number of trainable parameters during continual domain-specific pre-training of \fpdm\ in the Scientific Domain. LoRA is applied on the upper encoder of \fpdm\ during pre-training. \fpdm\ performs significantly better compared to when using LoRA in downstream NER, when there are a large number of classes (as in JNLPBA, which has $11$ classes), or when the dataset is extremely imbalanced (as in NCBI-Disease, where $91.72\%$ of the training samples belong to a single class). We attribute this to an insufficient number of trainable parameters when using LoRA during pre-training. Similarly, LoRA performs poorly in Relation and Text Classification. On the contrary, there is a meagre reduction in compute from 1.7 to 1.66 GPU-Hours when using $\fpdm(Sci.)_{BERT}(LoRA)$ instead of $\fpdm(Sci.)_{BERT}$, suggesting that LoRA is not beneficial from a compute perspective as well.

\subsection{Absence of document supervision}
\label{appendix:docsup}

\begin{table}[H]
\small
	\centering
\begin{tabular}{l|cc|c|c}
\hline
 \textbf{Dataset} & \scibert & \begin{tabular}[c]{@{}c@{}}\textbf{$\fpdm$}\\\textbf{(3 hier. levels)}\end{tabular}  & \textbf{\begin{tabular}[c]{@{}c@{}}$\fpdm$\\(w/o est. meta, tax. \\3 hier. levels)\end{tabular}} & \textbf{\begin{tabular}[c]{@{}c@{}}$\fpdm$\\(w/o est. meta, tax. \\11 hier. levels)\end{tabular}} \\ \hline
BC5CDR           & 85.55              & 87.81         & 87.6 & \textbf{87.88}          \\
JNLPBA & 59.5    & 75.84 & 75.91 & \textbf{76.06}\\
NCBI-D & \textbf{91.03}  &  84.33 & 85.02 & 85.05 \\
ChemProt  & 78.55    & \textbf{80.48} & 76.9 & 76.6 \\
      \hline
SciERC       & 74.3   & 78.95 & 79.26 & \textbf{81.21} \\
\hline

SciCite      & \textbf{84.44}  &  83.59  & 83.6 & 83.6 \\
\hline
\end{tabular}
	\caption{Results of \fpdm\ on tasks mentioned in \citet{scibert} in Scientific Domain with and without established domain-specific document metadata and taxonomy, compared to a well-performing baseline}
	\label{tab:scibert_paper_tasks_docsup}
\end{table}

\begin{table}[H]
\small
\centering
\begin{tabular}{c|cc}
\hline
\textbf{Model} & \multicolumn{1}{c}{\textbf{AUPR}}     &           \begin{tabular}[c]{@{}c@{}}\textbf{Precision@}\\ \textbf{80\% Recall}\end{tabular}                 \\ \hline
\rb + Contracts Pre-training & 45.2  & 34.1 \\
$\fpdm(Leg.)_{RoBERTa}$  (4 hier. levels)         & \begin{tabular}[c]{@{}c@{}}44.8\\(-0.88\%) \end{tabular} & \begin{tabular}[c]{@{}c@{}}34.6\\(+1.47\%) \end{tabular} \\ 
\hline
\begin{tabular}[c]{@{}c@{}}$\fpdm(Leg.)_{RoBERTa}$    \\(w/o est. meta., tax., 4 hier. levels)   \end{tabular}     & \begin{tabular}[c]{@{}c@{}}46.7\\ (+3.32\%)\end{tabular} & \begin{tabular}[c]{@{}c@{}}38.7 \\ (+13.49\%)\end{tabular} \\
\hline

\hline
\begin{tabular}[c]{@{}c@{}}$\fpdm(Leg.)_{RoBERTa}$    \\(w/o est. meta., tax., 17 hier. levels)   \end{tabular}     & \textbf{\begin{tabular}[c]{@{}c@{}}47.9\\ (+5.97\%)\end{tabular}} & \textbf{\begin{tabular}[c]{@{}c@{}}42 \\ (+23.17\%)\end{tabular}} \\
\hline
\end{tabular}%
\caption{Results of \fpdm\ on CUAD Dataset in Legal Domain with and without established domain-specific document metadata and taxonomy, compared to a well-performing baseline}
\label{tab:Leg_docsup}
\end{table}

\subsection{Why \fpdm\ works: Analysis of the interoperability of embeddings}

\label{FT_supp}

\noindent \textbf{Interoperability of pre-trained encoder parameters for input token and sentence embeddings}

We present experiments and observations to support the surprising interoperability of input embeddings, in response to the following research questions -

\emph{\underline{Q1. How does \fpdm\ learn local context?}}

\begin{figure}[H]
    \centering
    \includegraphics[width=0.5\textwidth]{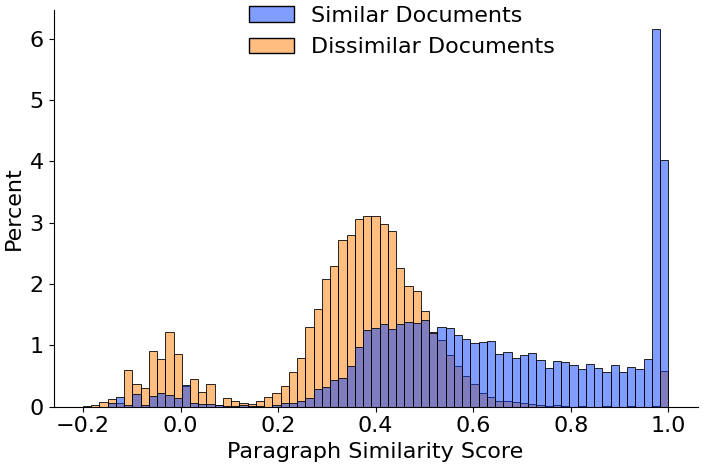}
    \caption{Distribution of `Paragraph Similarity Score' across similar and dissimilar document pairs}
    \label{fig:para_sim_1}
\end{figure}

\noindent\textbf{Nature of pre-training inputs:} 
We contrast the paragraph-level similarity between similar and dissimilar input documents used during pre-training. Given a pair of E-Manuals (from the Customer Support Domain), each paragraph in the two E-Manuals is converted to a fixed-size vector using Doc2Vec model \citep{doc2vec} trained on Wikipedia, and the similarity score (cosine) with the most similar paragraph from the other E-Manual is considered as a `Paragraph Similarity Score'. The distribution of this score across similar and dissimilar document pairs is plotted in Fig.~\ref{fig:para_sim_1}.

We can infer that paragraph pairs from similar documents are skewed towards higher similarity, with more than half of the samples having a `Paragraph Similarity Score' > $0.7$. The sections where the similarity is even higher are mainly the specific sections where some procedure/task is specified like `How to calibrate a wireless thermometer?' or the device specifications and/or warnings about using the device/service are given, which would indirectly help the model in the downstream QA task. Similarly, (even the most similar) paragraph pairs in dissimilar documents are skewed towards lower similarity, with a  majority of the samples having a `Paragraph Similarity Score' < $0.5$. Thus similar documents have very-similar local (paragraph-level) contexts, suggesting that, using document-level supervision during pre-training also helps in learning local contexts.

\textbf{Analysis of Local Context learned using \fpdm: }Document-level supervision using \fpdm~performs well on token-level tasks that require learning from local context (such as NER, relation classification, etc.), even though it was not explicitly trained on learning local context. To validate this, we measure the influence of local context on a random token vis-a-vis a standard MLM model (that learns from local context in accordance with the supervision signal). We randomly sample 500 sentences from each of the 3 domains. For each sentence, we take a random token and calculate the change in its prediction probability on masking other tokens in the sentence. Table \ref{tab:appendix_masked_token_prob} shows the Spearman Correlation of this change between two models (\fpdm~and a domain-specific model pre-trained using MLM) for each domain. We observe that the correlation is moderately high for all domains, showing that the local contexts is learned by \fpdm~to a reasonable extent.


\begin{table}[!thb]
\small
\centering
\begin{tabular}{c|c|c|c}
\hline
\textbf{Domain}                                             & \textbf{Model using} \fpdm & \textbf{Model using MLM} & \textbf{Correlation} \\ \hline
\begin{tabular}[c]{@{}c@{}}Customer\\ Support\end{tabular} & $\fpdm(Cus.)_{RoBERTa}$   & \emrb          & 0.368 \\
\begin{tabular}[c]{@{}c@{}}Scientific\\ Domain\end{tabular} & $\fpdm(Sci.)_{BERT}$                &    \scibert               & 0.481                \\
\begin{tabular}[c]{@{}c@{}}Legal\\ Domain\end{tabular}     & $\fpdm(Leg.)_{RoBERTa}$ & \begin{tabular}[c]{@{}c@{}}\rbe +\\Contracts Pre-training\end{tabular} & 0.393     \\ \hline
\end{tabular}%
\caption{Correlation of the change in masked token prediction probability between \fpdm~and MLM, corresponding to other masked tokens, across domains.}
\label{tab:appendix_masked_token_prob}
\end{table}

\emph{\ul{Q2. Are the relative representations preserved across the two embedding spaces? }}

\begin{figure}[!thb]
	\centering
	\subfloat[with sentence embeddings as inputs]{%
		\centering \includegraphics[width=0.8\columnwidth]{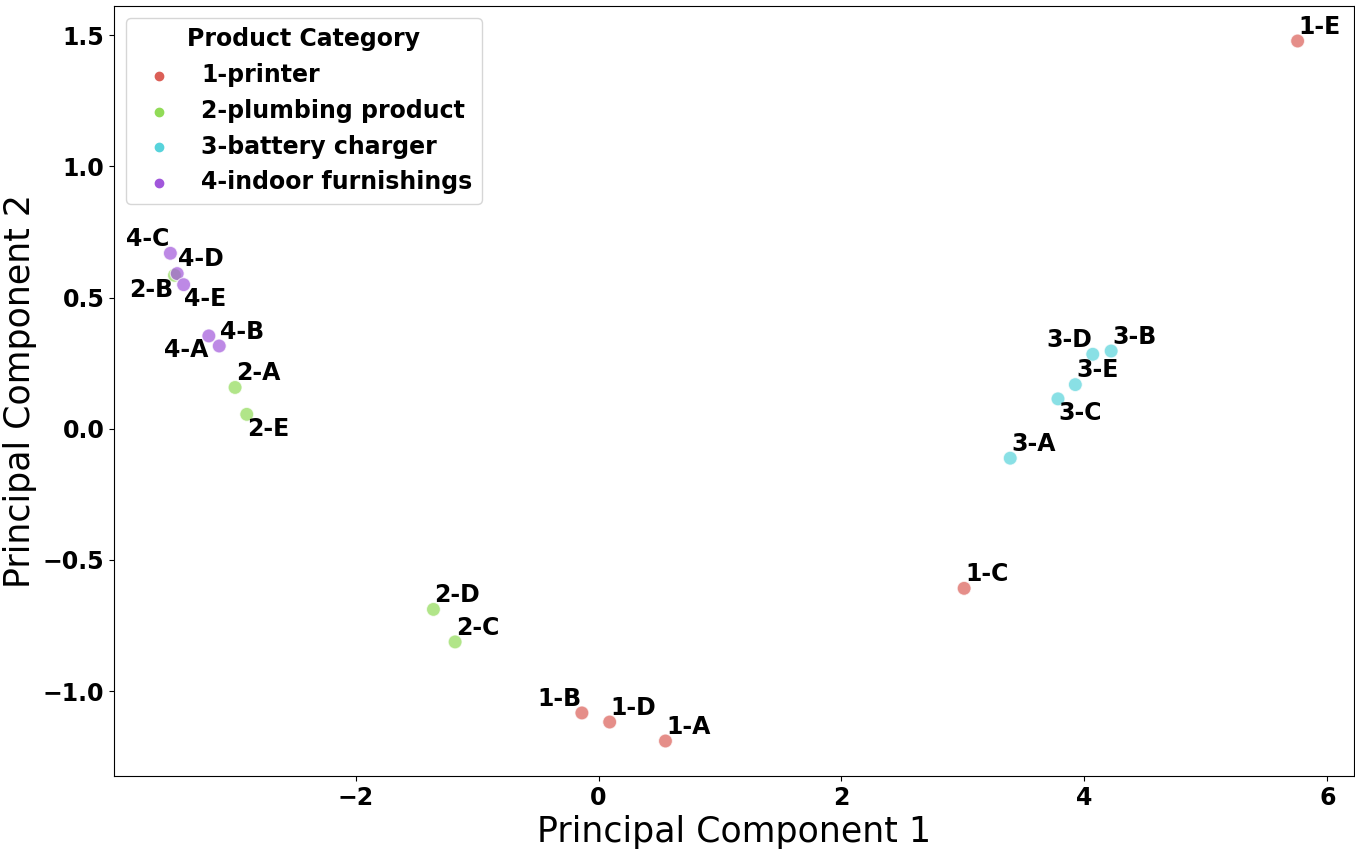} \label{roberta_sent}}\\
	\subfloat[with token embeddings as inputs]{%
		\centering \includegraphics[width=0.8\columnwidth]{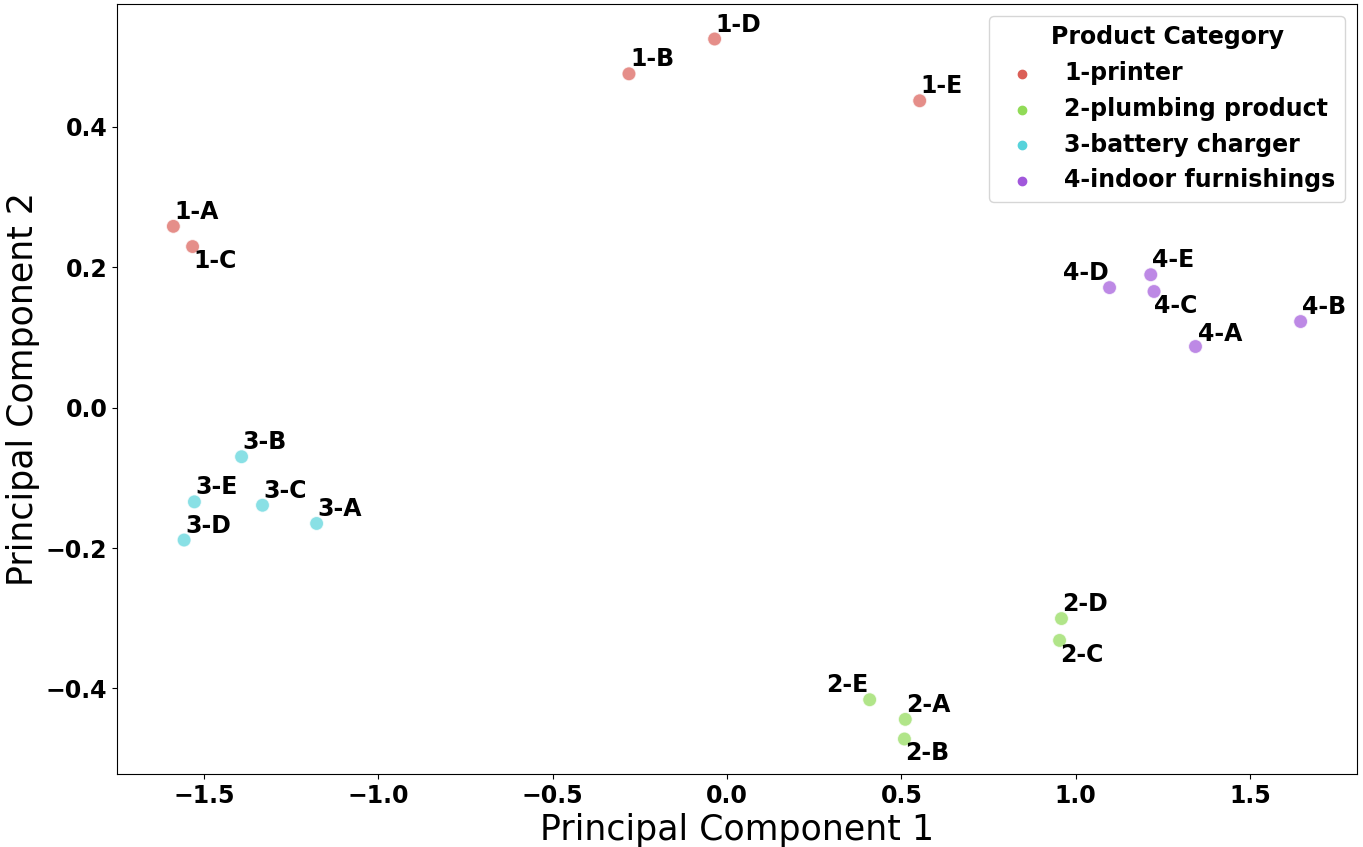} \label{roberta_tok}}
	\caption{2D Plots of first two principal components of the document representations of 20 E-Manuals from 4 product categories using $\fpdm\textit{(Cus.)}_{RoBERTa}$ for different types of input embeddings}
	\label{fig:doc_repr}
\end{figure}

\newcommand\Mycomb[2][^{20}]{\prescript{#1\mkern-0.5mu}{}C_{#2}}
\noindent\textbf{Qualitative Evaluation of  Relative Document Representations: }We analyze the relative document representations  learnt by the pre-trained encoder in $\fpdm\textit{(Cus.)}_{RoBERTa}$, for both sentence-level and token-level input embeddings. 
For $4$ different product categories (printer, plumbing product, battery charger, indoor furnishings), we consider $5$ E-Manuals each containing between $400-512$ tokens 
 so that it complies with the 
maximum number of tokens accepted by \bert or \rbe as inputs. 
The cosine similarity between the E-Manual representations of each of the $\Mycomb{2} = 190$ E-Manual pairs corresponding to both types of input embeddings is obtained, and normalized (using max-min normalization). The similarity values {corresponding to the two types of input embeddings} are positively correlated to each other, with the Pearson Correlation value being $0.515$. 
We further take the category-wise average representations, and repeat this experiment. We find that the Correlation for the similarity values is $0.977$. Hence, the relative representations in both the representation spaces are highly correlated, which justifies good downstream performance when an encoder pre-trained on sentence embedding inputs is fine-tuned on token embedding inputs.

For a visual analysis of these E-manuals, PCA (Principal Component Analysis) is applied over the document representations to reduce the vector dimension from $768$ to $2$. These representations are then plotted (as shown in Fig. \ref{fig:doc_repr}) for the pre-trained encoder in $\fpdm\textit{(Cus.)}_{RoBERTa}$ and two types of input embeddings (sentence and token level). Different product categories are shown in different colors. We 
infer that independent of whether the inputs are sentence or token-level embeddings, the E-Manuals are clustered in a similar manner across the two representation spaces, and hence, the relative representations are preserved across the two embedding spaces.

\emph{\ul{Q3. How are pre-training and fine-tuning compatible?}}

\textbf{Compatibility between pre-training using \fpdm~and downstream tasks via few-shot fine-tuning: }To test this compatibility, we fine-tune on a small number of samples in a few-shot setting. We perform 50-shot fine-tuning (i.e., fine-tuning on 50 training samples) on 3 tasks from 3 domains - Span-Based Question Answering on TechQA Dataset from Customer Support (with no intermediate SQuAD Fine-tuning), Text Classification on SciCite Dataset from Scientific Domain, and Span Extraction on CUAD Dataset from Legal Domain. We compared \fpdm~(pre-trained using document-level supervision) with \rbe/\bert (pre-trained without any document-level supervision). Tables \ref{tab:appendix_CS_few_shot}, \ref{tab:appendix_Sci_few_shot}, and \ref{tab:appendix_Leg_few_shot} show the results, along with the respective improvements when using \fpdm. Better performance of \fpdm~across all 3 domains in few-shot fine-tuning setting suggests that (1) Pre-training using document-level supervision is effective across 3 domains (2) \fpdm~Pre-training and Fine-tuning are compatible.

\begin{table}[!thb]
\small
\centering
\begin{tabular}{c|c|c|c}
\hline
\textbf{Model} &
  \textbf{F1} &
  \textbf{HA\_F1@1} &
  \textbf{HA\_F1@5} \\ \hline
\rbe &
  0.71 &
  1.38 &
  2.71 \\
$\fpdm(Cus.)_{RoBERTa}$ &
  \textbf{\begin{tabular}[c]{@{}c@{}}0.86\\ (+21.1\%)\end{tabular}} &
  \textbf{\begin{tabular}[c]{@{}c@{}}1.66\\ (+20.3\%)\end{tabular}} &
  \textbf{\begin{tabular}[c]{@{}c@{}}4.85\\ (+79\%)\end{tabular}} \\ \hline
\end{tabular}%
\caption{Results on TechQA Dataset in Customer Support Domain}
\label{tab:appendix_CS_few_shot}
\end{table}

\begin{table}[!thb]
\small
\centering
\begin{tabular}{c|c}
\hline
\textbf{Model} & \multicolumn{1}{c}{\textbf{Macro F1}}                            \\ \hline
\bert           & 37.75                                                             \\
$\fpdm(Sci.)_{BERT}$           & \textbf{\begin{tabular}[c]{@{}c@{}}40.16\\ (+6.4\%)\end{tabular}} \\ \hline
\end{tabular}%
\caption{Results on SciCite Dataset in Scientific Domain}
\label{tab:appendix_Sci_few_shot}
\end{table}

\begin{table}[!thb]
\small
\centering
\begin{tabular}{c|c}
\hline
\textbf{Model} & \multicolumn{1}{c}{\textbf{AUPR}}                                \\ \hline
\rbe        & 0.13                                                              \\
$\fpdm(Leg.)_{RoBERTa}$           & \textbf{\begin{tabular}[c]{@{}c@{}}0.14\\ (+7.69\%)\end{tabular}} \\ \hline
\end{tabular}%
\caption{Results on CUAD Dataset in Legal Domain}
\label{tab:appendix_Leg_few_shot}
\end{table}

\begin{figure}[!thb]
	\centering
	\captionsetup{justification=centering}
	\subfloat[Sentence Embeddings of similar pairs]{%
		\centering 
\includegraphics[width=0.5\textwidth]{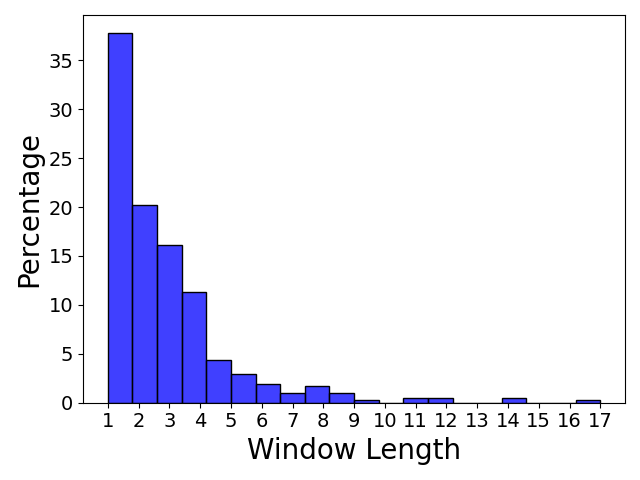} \label{sent_sim_wl}}
	\subfloat[Sentence Embeddings of dis-similar pairs]{%
		\centering 
\includegraphics[width=0.5\textwidth]{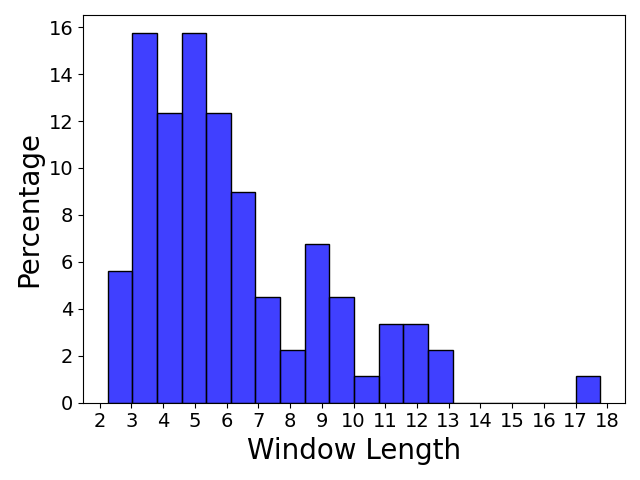} \label{sent_dissim_wl}}
		\\
	
	\subfloat[Token Embeddings of similar pairs]{%
		\centering 
\includegraphics[width=0.5\textwidth]{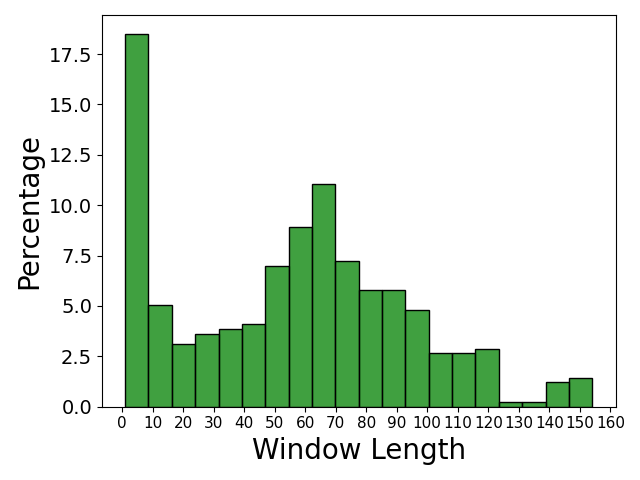} \label{tok_sim_wl}}
	\subfloat[Token Embeddings of dis-similar pairs]{%
		\centering 
\includegraphics[width=0.5\textwidth]{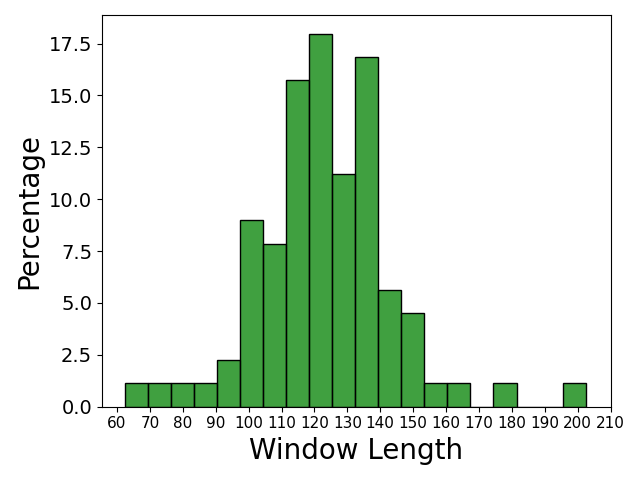} \label{tok_dissim_wl}}
\captionsetup{justification   = raggedright,
              singlelinecheck = false}
	\caption{Distribution of \wl~for sentence and token embeddings as inputs to pre-trained $\fpdm\textit{(Cus.)}_{RoBERTa}$ encoder for similar and dis-similar document pairs}
	\label{fig:wl}
\end{figure}


\noindent\emph{\ul{Additional Analysis}}

\noindent\textbf{Experiment on analyzing local context similarity of input embeddings: } Fig. \ref{fig:wl} shows the distribution of \wl~(Window Length) corresponding to input sentence and
token embeddings for
RoBERTa-based $\fpdm\textit{(Cus.)}_{RoBERTa}$ encoder for similar and dis-similar
document pairs, for documents that have between $400-512$ tokens. Given a pair of documents, the first being an anchor, \wl~is $1$ more than the distance between an input embedding of the first document, and the most similar input embedding of the second document, averaged across all embeddings of the first document.
If similar embeddings are present at nearby positions in two documents, \wl~ will tend to be smaller. Thus \wl~quantifies the local context similarity of the input embeddings. We observe that the distribution of \wl~is skewed towards smaller values (i.e., inputs are locally more similar) in similar pairs compared to dis-similar pairs, irrespective of the input (token or sentence) embeddings. Additionally, this also suggests that similar documents inherently induce learning of similarity between sentences and tokens when using \fpdm, thus learning from local contexts.

\end{document}